\documentclass[conference]{IEEEtran}
\IEEEoverridecommandlockouts

\usepackage{cite}
\usepackage{amsmath,amssymb,amsfonts}
\usepackage{algorithmic}
\usepackage{graphicx}
\usepackage{textcomp}
\usepackage{xcolor}

\usepackage{algorithm}
\usepackage{algorithmic}
\usepackage{subfigure}
\usepackage{lipsum}
\usepackage{bm}
\usepackage{amsfonts}
\usepackage{amsmath}
\usepackage{color, colortbl}
\usepackage{multirow}

\def\BibTeX{{\rm B\kern-.05em{\sc i\kern-.025em b}\kern-.08em
    T\kern-.1667em\lower.7ex\hbox{E}\kern-.125emX}}
\begin{document}

\title{Extending Cox Proportional Hazards Model with Symbolic Non-Linear Log-Risk Functions for Survival Analysis
}

\author{\IEEEauthorblockN{Jiaxiang Cheng}
\IEEEauthorblockA{\textit{School of Electrical and Electronic Engineering} \\
\textit{Nanyang Technological University}\\
Singapore \\
jiaxiang002@e.ntu.edu.sg}
\and
\IEEEauthorblockN{Guoqiang Hu}
\IEEEauthorblockA{\textit{School of Electrical and Electronic Engineering} \\
\textit{Nanyang Technological University}\\
Singapore \\
gqhu@ntu.edu.sg}
}

\maketitle

\begin{abstract}
The Cox proportional hazards (CPH) model has been widely applied in survival analysis to estimate relative risks across different subjects given multiple covariates. Traditional CPH models rely on a linear combination of covariates weighted with coefficients as the log-risk function, which imposes a strong and restrictive assumption, limiting generalization. Recent deep learning methods enable non-linear log-risk functions. However, they often lack interpretability due to the end-to-end training mechanisms. The implementation of Kolmogorov-Arnold Networks (KAN) offers new possibilities for extending the CPH model with fully transparent and symbolic non-linear log-risk functions. In this paper, we introduce Generalized Cox Proportional Hazards (GCPH) model, a novel method for survival analysis that leverages KAN to enable a non-linear mapping from covariates to survival outcomes in a fully symbolic manner. GCPH maintains the interpretability of traditional CPH models while allowing for estimation of non-linear log-risk functions. Experiments conducted on both synthetic data and public benchmarks demonstrate that GCPH achieves competitive performance in terms of prediction accuracy and exhibits superior interpretability compared to current state-of-the-art methods.
\end{abstract}


\section{Introduction}

Survival analysis is widely applied across various industries to predict survival probabilities and estimate risks over the lifetime of different subjects. One of the most popular models for this purpose is the Cox proportional hazards (CPH) model, which models the relationship between covariates of subjects and their survival outcomes~\cite{cox_regression_1972}. The \textit{hazard function} in the CPH model is specified as:
\begin{equation}
h(t \,|\, \mathbf{x}) = h_0(t) e^{\bm{\beta} \mathbf{x}},
\label{eq:hazard}
\end{equation}
where \( t \) is the time, \( \mathbf{x} \) is the covariate vector with coefficients \( \bm{\beta} \), and \( h_0(t) \) is the baseline hazard function, identical for all subjects. The \textit{hazard ratio}, or relative risk, between two subjects with covariates \( \mathbf{x}_1 \) and \( \mathbf{x}_2 \) is compared as:
\begin{equation}
\frac{h_1(t \,|\, \mathbf{x}_1)}{h_2(t \,|\, \mathbf{x}_2)} = \frac{h_0(t) e^{\bm{\beta} \mathbf{x}_1}}{h_0(t) e^{\bm{\beta} \mathbf{x}_2}} = \frac{e^{\bm{\beta} \mathbf{x}_1}}{e^{\bm{\beta} \mathbf{x}_2}}=e^{\bm{\beta} (\mathbf{x}_1-\mathbf{x}_2)},
\end{equation}
which is independent of time \( t \) and only related to the covariates. Therefore, the CPH model is commonly referred to as a \textit{semi-parametric model}, as it only requires partial parameters to be specified. The CPH model has been extensively applied due to its efficiency in evaluating relative risks between subjects with different covariates.
The term \( f(\mathbf{x}; \bm{\beta}) = \bm{\beta} \mathbf{x} \) in Equation \ref{eq:hazard} is also called the \textit{log-risk function}. While this linear assumption in the traditional CPH model simplifies the modeling process regarding parameter estimation, it also highly restricts the generalization ability when dealing with more complex and non-linear relationships. Consequently, recent research has focused on extending the CPH model to 
incorporate a \textit{non-linear log-risk function}.

\begin{figure}[!t]\centering
\subfigure[Ground Truth]{
\label{fig:non-linear-true}
\centering
\includegraphics[width=1.25in]{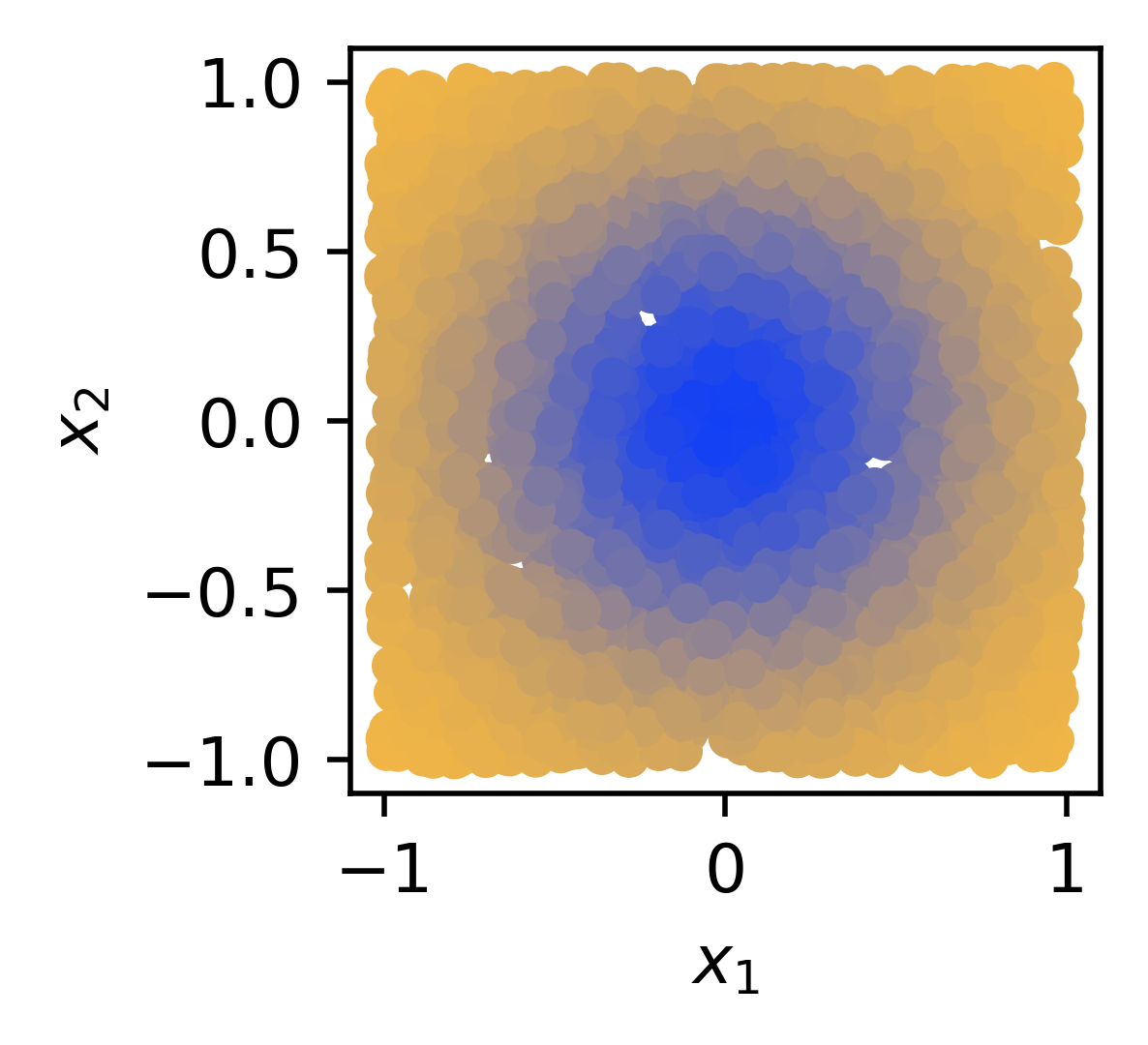}
}%
\hfil
\subfigure[CPH]{
\centering
\includegraphics[width=1.25in]{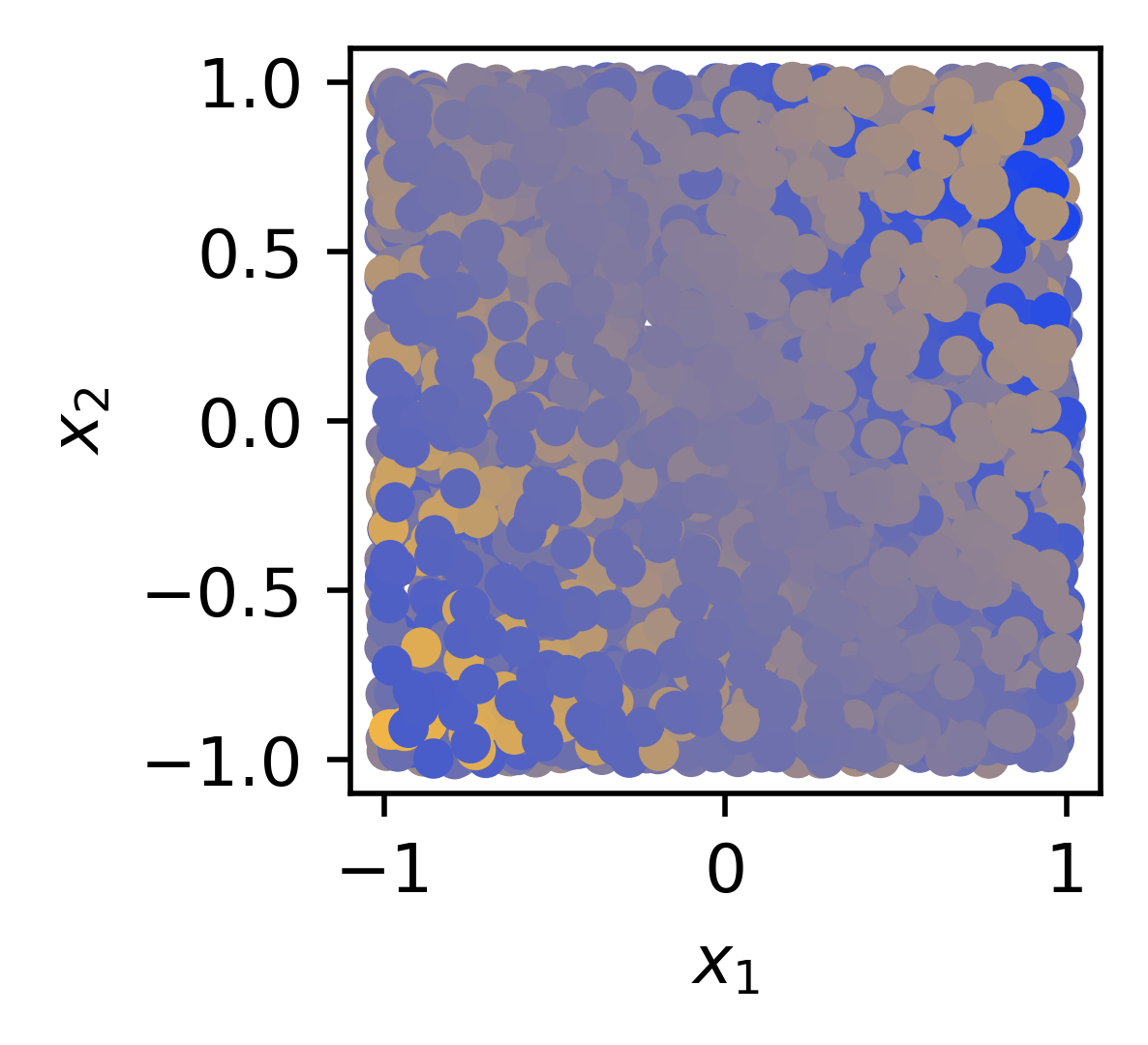}
}%
\hfil

\subfigure[DCPH]{
\centering
\includegraphics[width=1.25in]{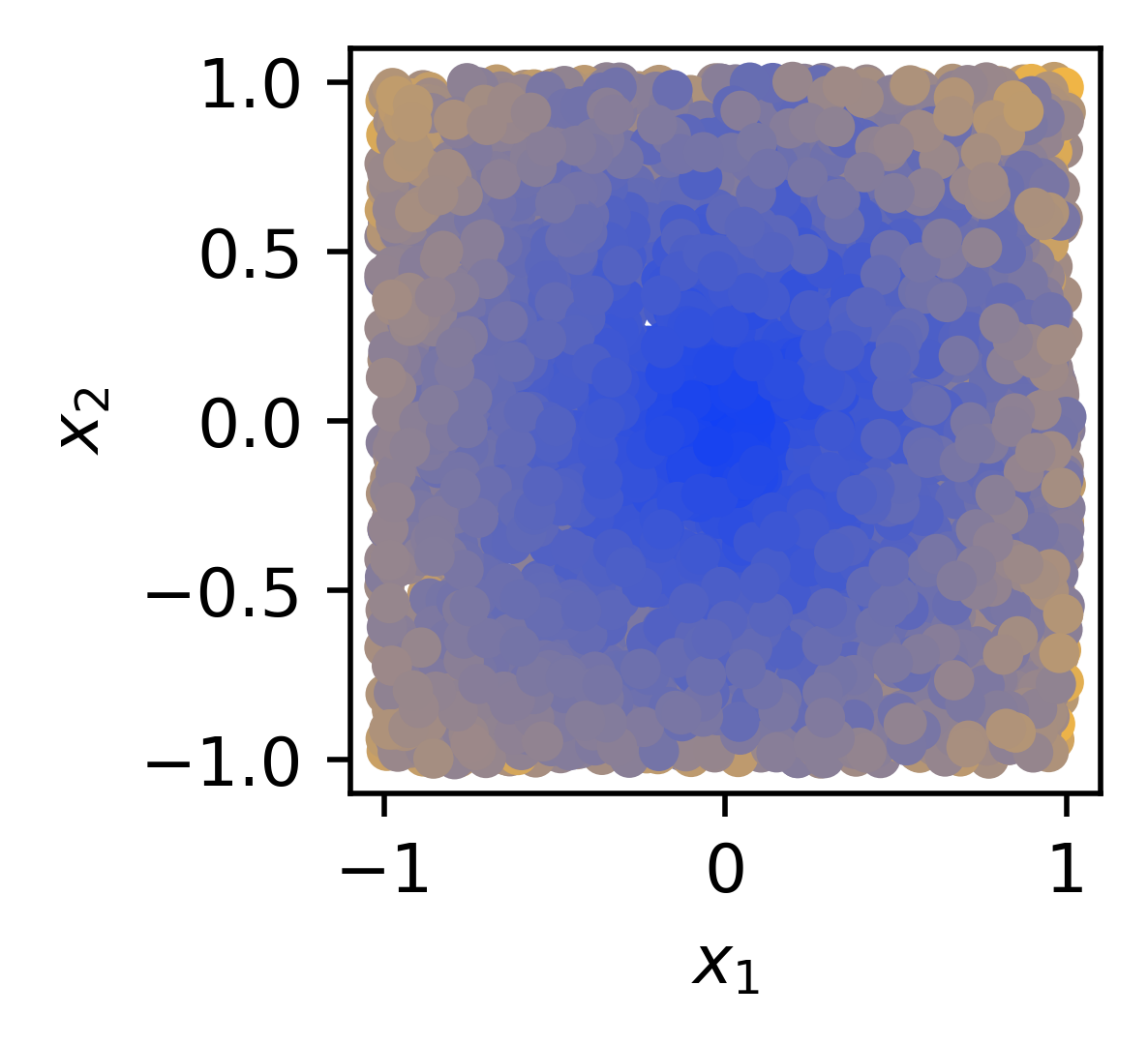}
}%
\hfil
\subfigure[\textbf{\textit{Proposed}}]{
\centering
\includegraphics[width=1.25in]{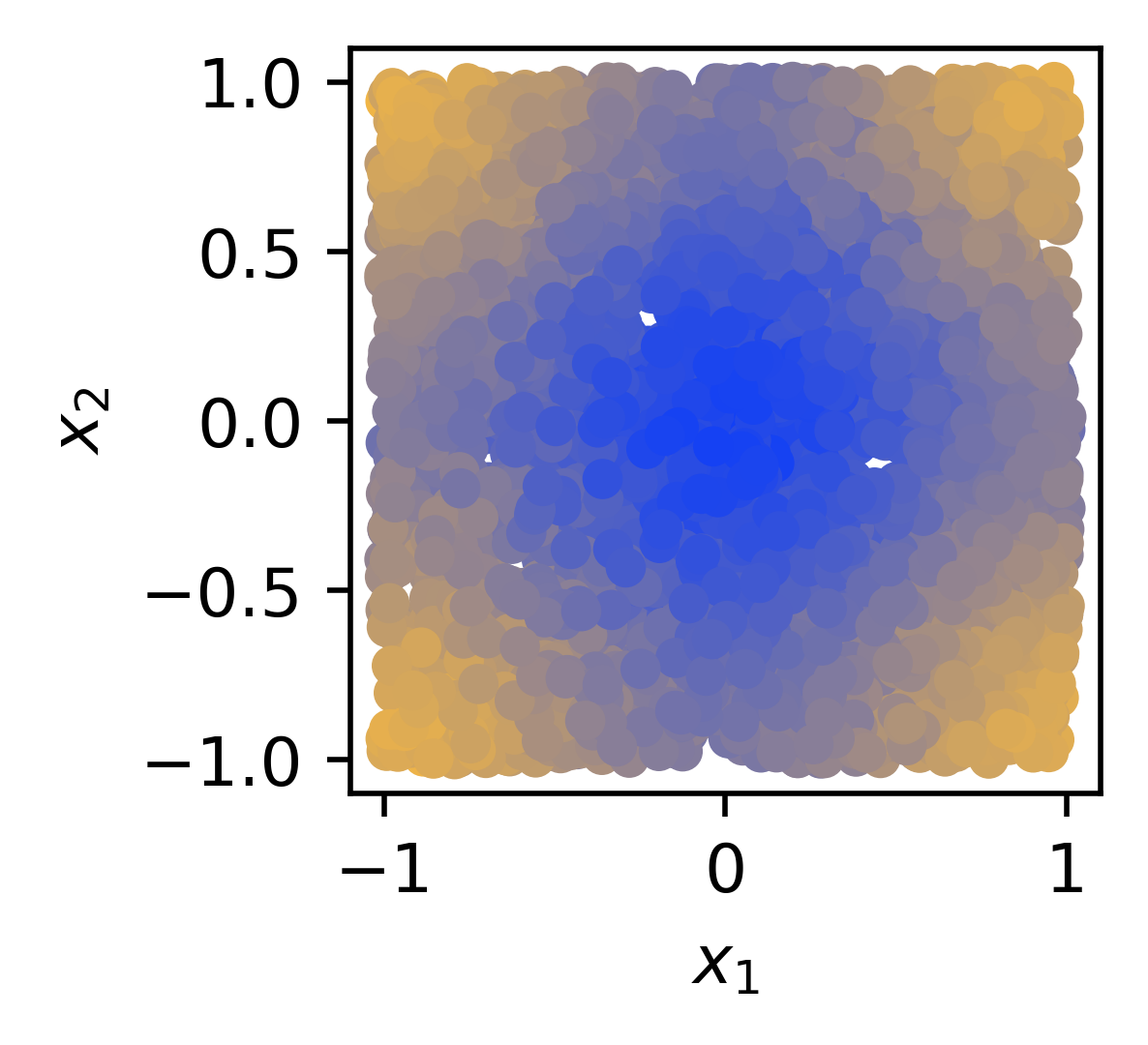}
}%
\caption{\textbf{Comparison of learning abilities from non-linear relationships}. The experiments are based on synthetic non-linear data, illustrated in (a). The models compared are (b) traditional CPH~\cite{cox_regression_1972}, (c) DCPH~\cite{katzman_deepsurv_2018}, and (d) our proposed model. \textit{This demonstrates the enhanced capability of proposed model in capturing non-linear relationships compared to existing methods.}}
\label{fig:nonlinear}
\end{figure}

Previous work~\cite{faraggi_neural_1995,katzman_deepsurv_2018} has proposed using neural networks to approximate the log-risk function as $f(\mathbf{x}; \bm{\theta})$, where $\bm{\theta}$ are the parameters optimized in the neural networks. Deep neural networks or multi-layer perceptrons (MLP) are effective in modeling the non-linear relationship between inputs and outputs. However, with increasing concerns about model interpretability, a more transparent survival model is needed to extend the traditional CPH model with generalized non-linearity.
Therefore, to overcome the challenges, the main contributions of this paper can be summarized as follows:
\begin{itemize}
    \item To the best of our knowledge, this is at least one of the first efforts to achieve fully symbolic derivation of the non-linear log-risk function in survival analysis by leveraging the Kolmogorov-Arnold Networks (KAN). This work was completed in Aug 19, 2024, with all experiments and results finalized in a public GitHub repository by the end of the same month\footnote{https://github.com/jiaxiang-cheng/KAN-for-Survival-Analysis}. Subsequently, a related preprint was published on Sep 6, 2024~\cite{knottenbelt_coxkan_2024}, which shares a similar model name and approach. And our study provides a complementary perspective.
    \item We propose a novel method for training a single-layer KAN model through a specialized loss function to approximate the log-risk function and predict hazards for different subjects in the context of survival analysis.
    \item Extensive experiments are conducted on various synthetic and public benchmarks, which are complementary to experiments by~\cite{knottenbelt_coxkan_2024}, to demonstrate the outstanding performance of our proposed model. The estimated symbolic functions provide new insights into understanding the effects of different variables on survival outcomes.
\end{itemize}

\section{Related Work}

\subsubsection{Machine Learning for Survival Analysis}
Machine learning (ML) has gained significant attention in recent years for survival analysis~\cite{wang_machine_2019}. One of the first successful ML-based models for survival analysis is the random survival forests (RSF) method. It handles censored data by introducing new survival splitting rules to the conventional random forests method~\cite{ishwaran_random_2008}, where the \textit{censored data} are the subjects with events not observed during study. The relevance vector machine was extended to improve computational efficiency and sparsity, thereby learning the nonlinear impacts of covariates on survival outcomes~\cite{kiaee_relevance_2016}. The deep multi-task Gaussian process (DMGP) was also utilized to model the relationships between input covariates and survival times~\cite{alaa_deep_2017}. 
In recent years, deep learning-based models have been widely proposed to handle large-scale data due to their outstanding ability to learn complex relationships among covariates. DeepHit models the \textit{time to event} as the hitting time in a stochastic process, rather than modeling it in continuous-time space~\cite{lee_deephit_2018}, whereas Nnet-survival models survival times in discrete-time space~\cite{gensheimer_scalable_2019}. Additionally, graph convolutional networks have also been applied to survival analysis to capture local neighbors from high-dimensional inputs~\cite{ling_survival_2022}.

\subsubsection{Extended Cox with Modified Log-Risk Function}
The CPH model continues to receive significant attention and has been extended for multi-tasking~\cite{liu_asymmetric_2022} and incorporating multi-modal data~\cite{ning_multi-constraint_2023}. Following the original practice of using neural networks for modeling survival data in~\cite{faraggi_neural_1995}, both DCPH (DeepSurv) and Cox-nnet combine modern deep neural networks with the inference mechanism of the CPH model to extract the impacts of different features on the hazard ratio~\cite{katzman_deepsurv_2018,ching_cox-nnet_2018}.
However, deep learning models often lack transparency and interpretability, which are major concerns in critical scenarios. The end-to-end training mechanisms pose challenges in uncovering the underlying principles of the relationships between predictors and response variables.

\section{Methodology}

In this section, we introduce the proposed model, which adapts KAN for approximating the log-risk function in survival analysis. We also present the specialized loss function that enables the prediction of survival outcomes.

\subsection{Model Architecture}

\begin{figure*}[t]
\centering
\includegraphics[width=2.03\columnwidth]{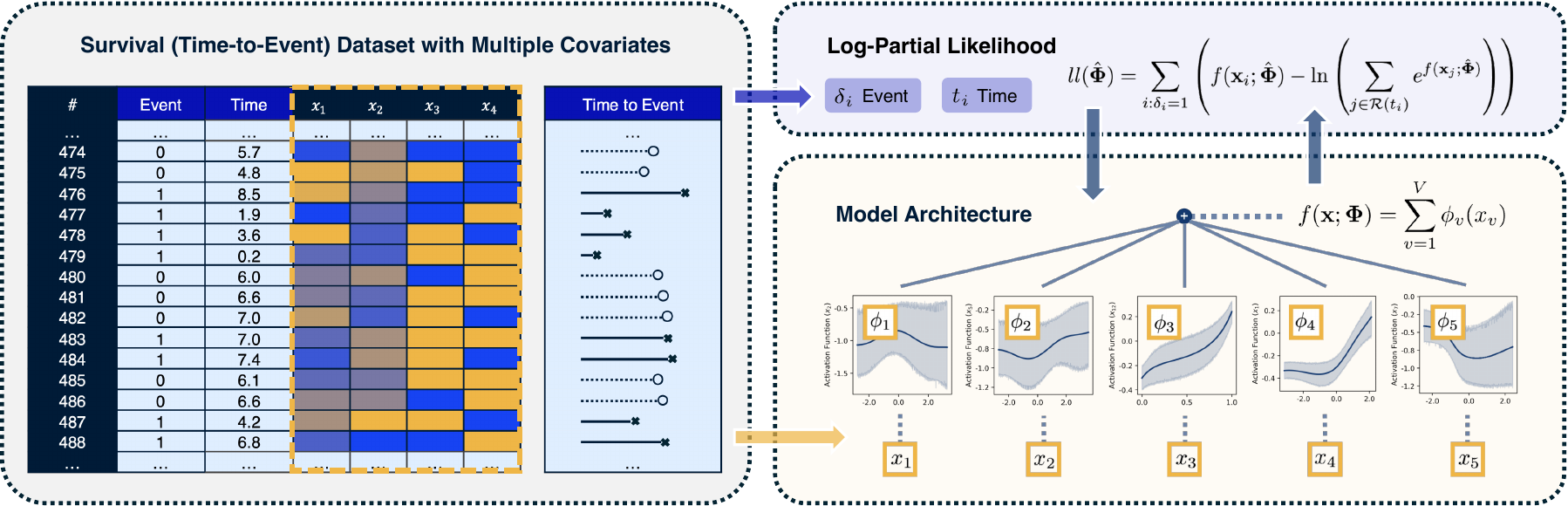} 
\caption{\textbf{Overview of proposed GCPH for survival analysis by extending CPH model with non-linear symbolic log-risk function}. A typical survival dataset is presented and used for training the GCPH with a specialized loss function, including the \textit{log-partial likelihood function}. Confidence interval is also illustrated for each \textit{symbolic activation function} through multiple tests with different random seeds for initialization.}
\label{fig1}
\end{figure*}

In this paper, instead of using deep learning models to approximate the log-risk function, we employ the KAN for symbolic approximation. Unlike deep neural networks, KAN directly estimates the linear or non-linear \textit{activation function} from each $v$-th feature $x_v$ to the log-risk as:
\begin{equation}
f(\mathbf{x; \bm{\Phi}}) = \sum^V_{v=1} \phi_v(x_v),
\end{equation}
which is equivalent to employing a single layer of KAN with a size of $V$ ($i.e.$, the number of covariates) as formulated in~\cite{liu_kan_2024}, with model architecture illustrated in Figure \ref{fig1}. For each covariate $x_v$, we approximate an independent function $\phi_v(\cdot)$, and the log-risk function is the summation of the outputs from all activations.

\subsubsection{Optimization}
To enable an optimizable activation function, each $\phi_v(\cdot)$ is defined as~\cite{liu_kan_2024}:
\begin{equation}
\phi_v(x_v) = \omega^b_v\, b(x_v) + \omega^s_v \,S_v(x_v),
\end{equation}
where $b(x)$ is a \textit{basis function} and $S_v(x)$ is the \textit{spline function}, defined and formulated as:
\begin{equation}
b(x_v) = \frac{x_v}{1 + e^{-x_v}}, 
\end{equation}
\begin{equation}
S_v(x_v) = \sum^K_{k=1} c_{v,k} \,B_{v,k}(x_v),
\label{eq:spline}
\end{equation}
respectively. Here, $S_v(x)$ is a linear combination of \textit{B-splines}, $i.e.$, $B_{v,k}(x_v)$, with $K$ the order. The scales of $b(x)$ and $S_v(x)$, $i.e.$, $\omega^b_v$ and $\omega^s_v$, and scales of $B_{v,k}(x_v)$, $i.e.$, $c_{v,k}$, are the trainable parameters.

\subsubsection{Symbolification}
To enable an explainable and transparent non-linear log-risk function, we employ the \textit{symbolification} process introduced in~\cite{liu_kan_2024}.
The process is straightforward: after the optimization process, we use several candidate symbolic functions $y(x)$ to approximate the optimized $\hat{\phi}_v(x)$ in the form:
\begin{equation}
\hat{\phi}_v(x_v) \approx \alpha_3 \, y(\alpha_1 x_v + \alpha_2) + \alpha_4,
\end{equation}
where $(\alpha_1, \alpha_2, \alpha_3, \alpha_4)$ are \textit{affine parameters} to be fitted. The optimal symbolic function $y^*_v(x)$ is then selected based on the best fitting performance, evaluated using $R^2$.

\subsection{Loss Function}
A specialized loss function is proposed that enables the KAN to approximate the non-linear log-risk function, which consists of log-partial likelihood function aligned with CPH model and regularization loss for preventing overfitting.

\subsubsection{Log-Partial Likelihood}
In the CPH model, the \textit{partial likelihood function} is constructed to optimize the coefficients $\bm{\beta}$ in Equation \ref{eq:hazard}, formulated as follows:
\begin{equation}
l(\hat{\bm{\beta}}) = \prod_{i:\delta_i=1} \frac{e^{f(\mathbf{x}_i; \hat{\bm{\beta}})}}{\sum_{j \in \mathcal{R}(t_i)} e^{f(\mathbf{x}_j; \hat{\bm{\beta}})}},
\end{equation}
where the product is taken over each $i$-th subjects where the event has occurred ($\delta_i=1$) and $\mathcal{R}(t_i)$ denotes the set of subjects still at risk at time $t_i$, $i.e.$, without events occurred by the time $t_i$. This function represents the probability of the event occurring for the $i$-th subject, given the total number of subjects still at risk at time $t_i$ ($i.e.$, the \textit{survival time} of the $i$-th subject). The regression is performed by maximizing the partial likelihood function $l(\hat{\bm{\beta}})$.

Similarly, for training the KAN model, we define the \textit{log-partial likelihood function} given the estimated $\hat{\bm{\Phi}}$ as:
\begin{equation}
ll(\hat{\bm{\Phi}}) = \sum_{i:\delta_i=1} \left( f(\mathbf{x}_i; \hat{\bm{\Phi}}) - \ln \left( \sum_{j \in \mathcal{R}(t_i)} e^{f(\mathbf{x}_j; \hat{\bm{\Phi}})} \right) \right)
\end{equation}

\subsubsection{Regularization Loss}
Regularization is applied as developed in~\cite{liu_kan_2024} to encourage sparsity and prevent overfitting of the KAN. Given an input batch $\mathcal{B}$, for each activation function $\phi_v$, the $L1$ norm is calculated as:
\begin{equation}
||\hat{\phi}_v||_1 = \frac{1}{|\mathcal{B}|} \sum_{x_v \text{ in } \mathbf{x} \in \mathcal{B}} |\hat{\phi}_v(x_v)|, \,\,\text{and}
\end{equation}
\begin{equation}
||\hat{\bm{\Phi}}||_1 = \sum_{v=1}^{V} ||\hat{\phi}_v||_1,
\end{equation}
where the $L1$ norm of the estimated $\hat{\bm{\Phi}}$ is calculated as the summation of the $L1$ norms of all activation functions.

Additionally, an \textit{entropy loss} is also introduced in~\cite{liu_kan_2024} and formulated as follows:
\begin{equation}
H(\hat{\bm{\Phi}}) = - \sum_{v=1}^{V} \frac{||\hat{\phi}_v||_1}{||\hat{\bm{\Phi}}||_1} \ln \left( \frac{||\hat{\phi}_v||_1}{||\hat{\bm{\Phi}}||_1} \right).
\end{equation}
The \textit{regularization loss} is the summation of the $L1$ norm of $\hat{\bm{\Phi}}$ and the entropy regularization loss:
\begin{equation}
\mathcal{L}_{\text{reg}}(\hat{\bm{\Phi}}) = \mu_1 ||\hat{\bm{\Phi}}||_1 + \mu_2 H(\hat{\bm{\Phi}}),
\label{eq:reg}
\end{equation}
where $\mu_1$ and $\mu_2$ are the weights assigned to the $L1$ norm and entropy regularization loss, respectively.

Thus, the total loss function is a combination of the \textit{negative log-partial likelihood} and the regularization loss:
\begin{equation}
\mathcal{L} = -ll(\hat{\bm{\Phi}}) + \gamma \, \mathcal{L}_{\text{reg}}(\hat{\bm{\Phi}}),
\label{eq:loss}
\end{equation}
where $\gamma$ is the weight assigned to the regularization loss.

\begin{table*}[!t]
\centering\fontsize{9}{11}\selectfont
\caption{\textbf{Summary of experimental results comparing proposed method with different baseline models}. C-Index and Brier Score computed with the predictions by different models on various benchmark.}
\begin{tabular}{>{\centering}p{1.6cm}>{\centering}p{1.7cm}|ccc|ccc}
\hline
 &  &   & \textbf{C-Index} $\uparrow$ & &  & \textbf{Brier Score} $\downarrow$ &  \\
   \cline{3-8}
Data & Model & 25$\%$ & 50$\%$ & 75$\%$ & 25$\%$ & 50$\%$ & 75$\%$ \\
\hline
\multirow{7}{*}{\shortstack{Synthetic \\ Linear}} & CPH & \textit{0.795} $_{(0.013)}$ & \textit{0.787} $_{(0.013)}$ & \textit{0.779} $_{(0.011)}$ & \textbf{0.128} $_{(0.007)}$ & \textit{0.150} $_{(0.011)}$ & \textit{0.124} $_{(0.010)}$ \\ 
 & RSF & 0.762 $_{(0.014)}$ & 0.770 $_{(0.013)}$ & 0.761 $_{(0.011)}$ & 0.141 $_{(0.007)}$ & 0.162 $_{(0.011)}$ & 0.133 $_{(0.013)}$ \\ 
& DCPH & 0.794 $_{(0.013)}$ & 0.786 $_{(0.012)}$ & 0.778 $_{(0.010)}$ & \textit{0.128} $_{(0.006)}$ & 0.151 $_{(0.011)}$ & 0.125 $_{(0.011)}$ \\ 
  & DCM & 0.793 $_{(0.013)}$ & 0.786 $_{(0.012)}$ & 0.778 $_{(0.011)}$ & 0.128 $_{(0.006)}$ & 0.151 $_{(0.011)}$ & 0.125 $_{(0.010)}$ \\ 
 & DSM & 0.792 $_{(0.012)}$ & 0.785 $_{(0.014)}$ & 0.776 $_{(0.012)}$ & 0.128 $_{(0.006)}$ & 0.151 $_{(0.011)}$ & 0.128 $_{(0.012)}$ \\ 
 \cline{2-8}
 & \textbf{GCPH} & 0.793 $_{(0.015)}$ & 0.783 $_{(0.012)}$ & 0.774 $_{(0.010)}$ & 0.129 $_{(0.008)}$ & 0.153 $_{(0.013)}$ & 0.129 $_{(0.011)}$ \\ 
 & \textbf{GCPH-$l$} & \textbf{0.796} $_{(0.013)}$ & \textbf{0.788} $_{(0.013)}$ & \textbf{0.779} $_{(0.011)}$ & 0.128 $_{(0.007)}$ & \textbf{0.150} $_{(0.012)}$ & \textbf{0.124} $_{(0.011)}$ \\ 
\hline
\multirow{6}{*}{\shortstack{Synthetic \\ Non-Linear}}  & CPH & 0.496 $_{(0.030)}$ & 0.500 $_{(0.022)}$ & 0.501 $_{(0.020)}$ & 0.176 $_{(0.011)}$ & 0.248 $_{(0.003)}$ & 0.220 $_{(0.006)}$ \\ 
 & RSF & 0.587 $_{(0.023)}$ & 0.582 $_{(0.023)}$ & 0.584 $_{(0.017)}$ & 0.178 $_{(0.012)}$ & 0.250 $_{(0.010)}$ & 0.221 $_{(0.009)}$ \\ 
 & DCPH & \textbf{0.626} $_{(0.030)}$ & \textbf{0.619} $_{(0.018)}$ & \textbf{0.616} $_{(0.016)}$ & \textit{0.168} $_{(0.011)}$ & \textbf{0.229} $_{(0.004)}$ & \textbf{0.200} $_{(0.007)}$ \\ 
 & DCM & 0.620 $_{(0.027)}$ & 0.614 $_{(0.016)}$ & \textit{0.612} $_{(0.016)}$ & 0.171 $_{(0.011)}$ & 0.236 $_{(0.005)}$ & 0.207 $_{(0.009)}$ \\ 
 & DSM & 0.560 $_{(0.045)}$ & 0.554 $_{(0.060)}$ & 0.550 $_{(0.062)}$ & 0.175 $_{(0.012)}$ & 0.247 $_{(0.007)}$ & 0.218 $_{(0.006)}$ \\ 
 \cline{2-8}
 & \textbf{GCPH} & \textit{0.624} $_{(0.028)}$ & \textit{0.616} $_{(0.018)}$ & 0.611 $_{(0.016)}$ & \textbf{0.167} $_{(0.013)}$ & \textit{0.230} $_{(0.005)}$ & \textit{0.202} $_{(0.009)}$ \\ 
\hline
\multirow{6}{*}{TRACE} & CPH & 0.757 $_{(0.031)}$ & 0.742 $_{(0.020)}$ & 0.738 $_{(0.013)}$ & \textit{0.093} $_{(0.008)}$ & 0.155 $_{(0.010)}$ & 0.180 $_{(0.007)}$ \\ 
 & RSF & 0.737 $_{(0.030)}$ & 0.738 $_{(0.017)}$ & 0.728 $_{(0.008)}$ & 0.096 $_{(0.008)}$ & 0.158 $_{(0.010)}$ & 0.186 $_{(0.006)}$ \\ 
 & DCPH & \textbf{0.762} $_{(0.029)}$ & \textbf{0.749} $_{(0.020)}$ & \textbf{0.742} $_{(0.013)}$ & \textbf{0.092} $_{(0.008)}$ & \textbf{0.152} $_{(0.011)}$ & \textbf{0.178} $_{(0.007)}$ \\ 
 & DCM & 0.753 $_{(0.034)}$ & 0.744 $_{(0.019)}$ & 0.737 $_{(0.011)}$ & 0.094 $_{(0.008)}$ & \textit{0.153} $_{(0.010)}$ & 0.180 $_{(0.004)}$ \\ 
 & DSM & 0.753 $_{(0.030)}$ & 0.738 $_{(0.018)}$ & 0.731 $_{(0.012)}$ & 0.094 $_{(0.009)}$ & 0.159 $_{(0.008)}$ & 0.185 $_{(0.006)}$ \\ 
 \cline{2-8}
 & \textbf{GCPH} & \textit{0.761} $_{(0.029)}$ & \textit{0.747} $_{(0.019)}$ & \textit{0.741} $_{(0.013)}$ & 0.093 $_{(0.008)}$ & 0.154 $_{(0.010)}$ & \textit{0.179} $_{(0.007)}$ \\ 
\hline
\multirow{6}{*}{COLON} & CPH & 0.700 $_{(0.022)}$ & \textbf{0.670} $_{(0.036)}$ & \textit{0.663} $_{(0.036)}$ & \textit{0.106} $_{(0.014)}$ & \textbf{0.171} $_{(0.016)}$ & \textit{0.206} $_{(0.016)}$ \\ 
 & RSF & 0.668 $_{(0.025)}$ & 0.651 $_{(0.026)}$ & 0.657 $_{(0.030)}$ & 0.109 $_{(0.013)}$ & 0.176 $_{(0.015)}$ & 0.208 $_{(0.016)}$ \\ 
 & DCPH & 0.676 $_{(0.031)}$ & 0.650 $_{(0.033)}$ & 0.649 $_{(0.033)}$ & 0.108 $_{(0.013)}$ & 0.175 $_{(0.015)}$ & 0.211 $_{(0.013)}$ \\ 
 & DCM & 0.683 $_{(0.027)}$ & 0.649 $_{(0.040)}$ & 0.653 $_{(0.039)}$ & 0.111 $_{(0.014)}$ & 0.178 $_{(0.015)}$ & 0.218 $_{(0.012)}$ \\ 
 & DSM & \textbf{0.704} $_{(0.024)}$ & 0.667 $_{(0.032)}$ & 0.660 $_{(0.036)}$ & 0.107 $_{(0.014)}$ & 0.175 $_{(0.018)}$ & 0.212 $_{(0.017)}$ \\ 
 \cline{2-8}
 & \textbf{GCPH} & \textit{0.703} $_{(0.028)}$ & \textit{0.668} $_{(0.039)}$ & \textbf{0.666} $_{(0.036)}$ & \textbf{0.106} $_{(0.013)}$ & \textit{0.171} $_{(0.016)}$ & \textbf{0.205} $_{(0.016)}$ \\ 
\hline
\multirow{6}{*}{RDATA} & CPH & \textbf{0.668} $_{(0.035)}$ & \textit{0.673} $_{(0.026)}$ & 0.674 $_{(0.020)}$ & 0.103 $_{(0.015)}$ & 0.177 $_{(0.014)}$ & \textbf{0.205} $_{(0.012)}$ \\ 
 & RSF & 0.644 $_{(0.020)}$ & 0.647 $_{(0.025)}$ & 0.653 $_{(0.025)}$ & 0.109 $_{(0.014)}$ & 0.187 $_{(0.018)}$ & 0.218 $_{(0.016)}$ \\ 
 & DCPH & 0.666 $_{(0.032)}$ & 0.672 $_{(0.028)}$ & \textit{0.674} $_{(0.021)}$ & \textit{0.103} $_{(0.015)}$ & \textit{0.176} $_{(0.014)}$ & 0.206 $_{(0.011)}$ \\ 
 & DCM & 0.661 $_{(0.035)}$ & 0.668 $_{(0.029)}$ & 0.670 $_{(0.023)}$ & 0.103 $_{(0.016)}$ & 0.179 $_{(0.014)}$ & 0.209 $_{(0.010)}$ \\ 
 & DSM & 0.663 $_{(0.037)}$ & 0.667 $_{(0.027)}$ & 0.669 $_{(0.023)}$ & \textbf{0.102} $_{(0.016)}$ & 0.177 $_{(0.015)}$ & 0.207 $_{(0.013)}$ \\ 
 \cline{2-8}
 & \textbf{GCPH} & \textit{0.666} $_{(0.029)}$ & \textbf{0.674} $_{(0.027)}$ & \textbf{0.676} $_{(0.023)}$ & 0.103 $_{(0.014)}$ & \textbf{0.176} $_{(0.014)}$ & \textit{0.206} $_{(0.013)}$ \\ 
\hline
\multirow{6}{*}{FRTCS} & CPH & \textbf{0.727} $_{(0.131)}$ & \textit{0.695} $_{(0.119)}$ & \textit{0.704} $_{(0.107)}$ & 0.025 $_{(0.011)}$ & 0.052 $_{(0.017)}$ & 0.074 $_{(0.022)}$ \\ 
 & RSF & 0.461 $_{(0.130)}$ & 0.625 $_{(0.095)}$ & 0.651 $_{(0.088)}$ & 0.026 $_{(0.011)}$ & 0.052 $_{(0.016)}$ & 0.072 $_{(0.019)}$ \\ 
 & DCPH & 0.644 $_{(0.169)}$ & 0.672 $_{(0.139)}$ & 0.700 $_{(0.109)}$ & 0.025 $_{(0.011)}$ & 0.052 $_{(0.017)}$ & \textit{0.072} $_{(0.022)}$ \\ 
 & DCM & 0.656 $_{(0.159)}$ & 0.671 $_{(0.103)}$ & 0.628 $_{(0.079)}$ & \textit{0.025} $_{(0.011)}$ & 0.051 $_{(0.016)}$ & 0.072 $_{(0.019)}$ \\ 
 & DSM & 0.677 $_{(0.151)}$ & 0.672 $_{(0.144)}$ & 0.664 $_{(0.132)}$ & 0.025 $_{(0.011)}$ & \textit{0.051} $_{(0.016)}$ & 0.072 $_{(0.020)}$ \\ 
 \cline{2-8}
 & \textbf{GCPH} & \textit{0.711} $_{(0.178)}$ & \textbf{0.718} $_{(0.106)}$ & \textbf{0.727} $_{(0.081)}$ & \textbf{0.020} $_{(0.005)}$ & \textbf{0.047} $_{(0.016)}$ & \textbf{0.069} $_{(0.021)}$ \\ 
\hline
\end{tabular}
\label{tab:summary}
\end{table*}

\begin{figure*}[!t]\centering
\subfigure[Synthetic Linear]{
\centering
\includegraphics[width=2.2in]{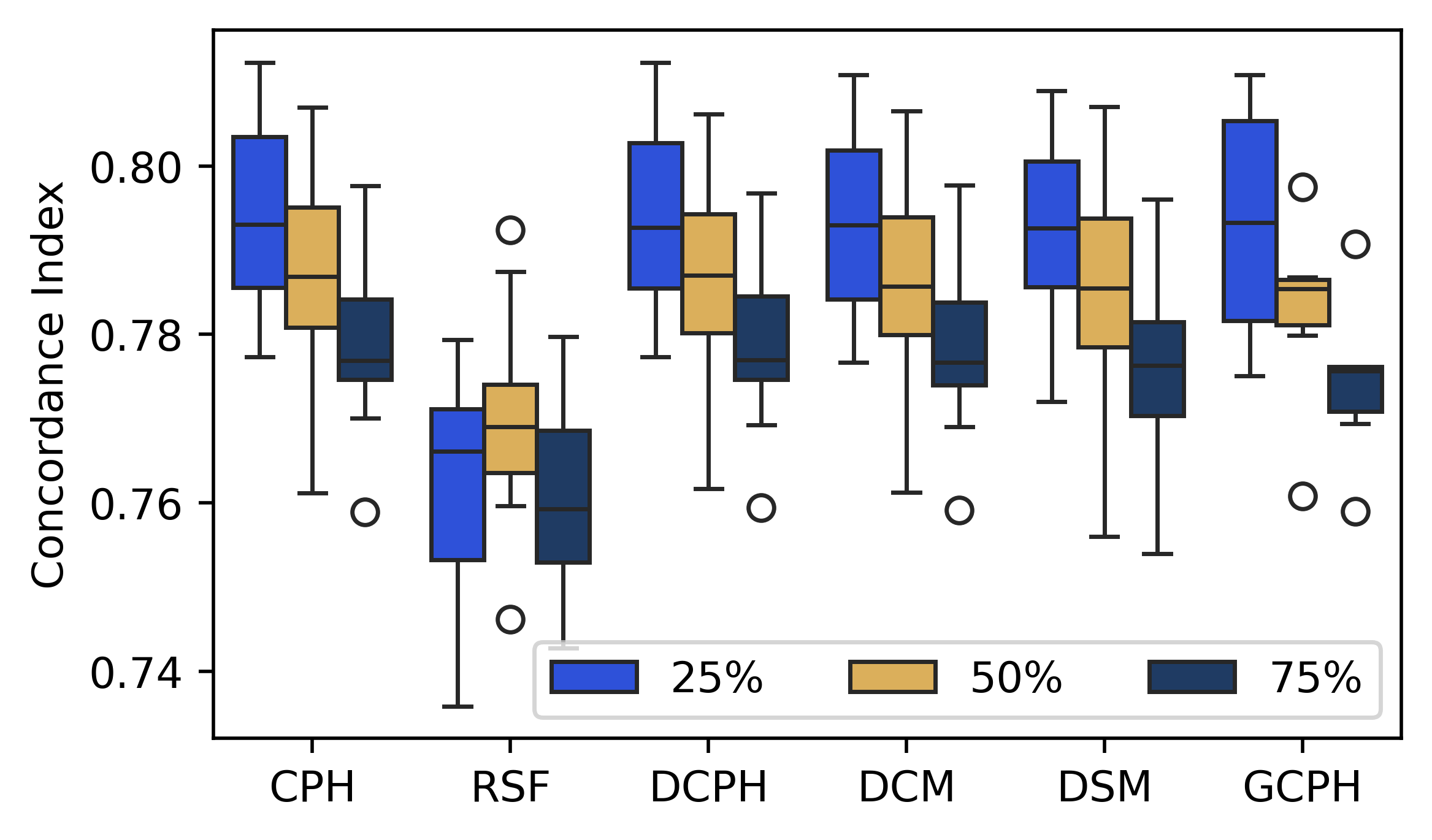}
}%
\hfil
\subfigure[Synthetic Non-Linear]{
\centering
\includegraphics[width=2.2in]{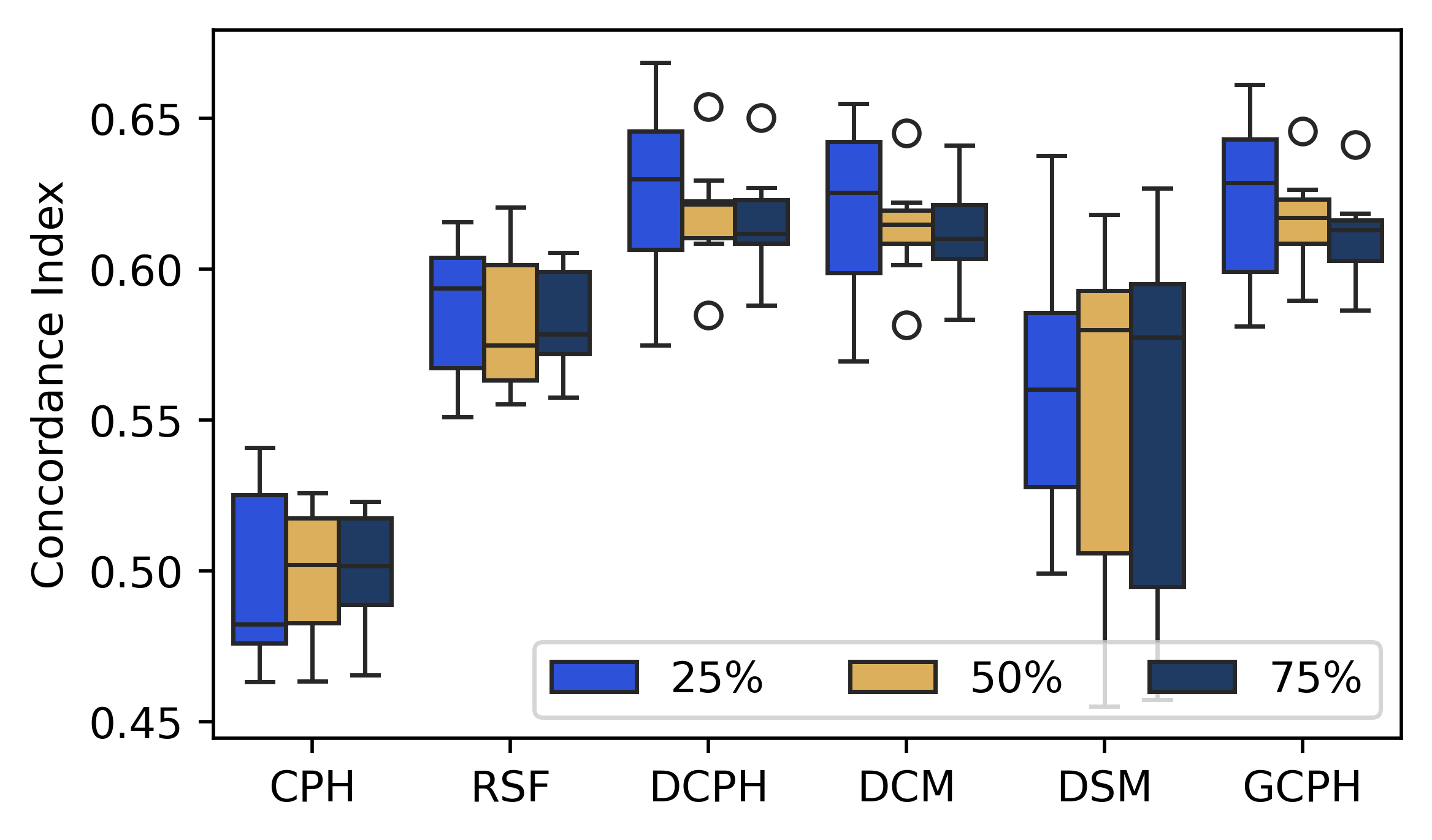}
}%
\hfil
\subfigure[TRACE]{
\centering
\includegraphics[width=2.2in]{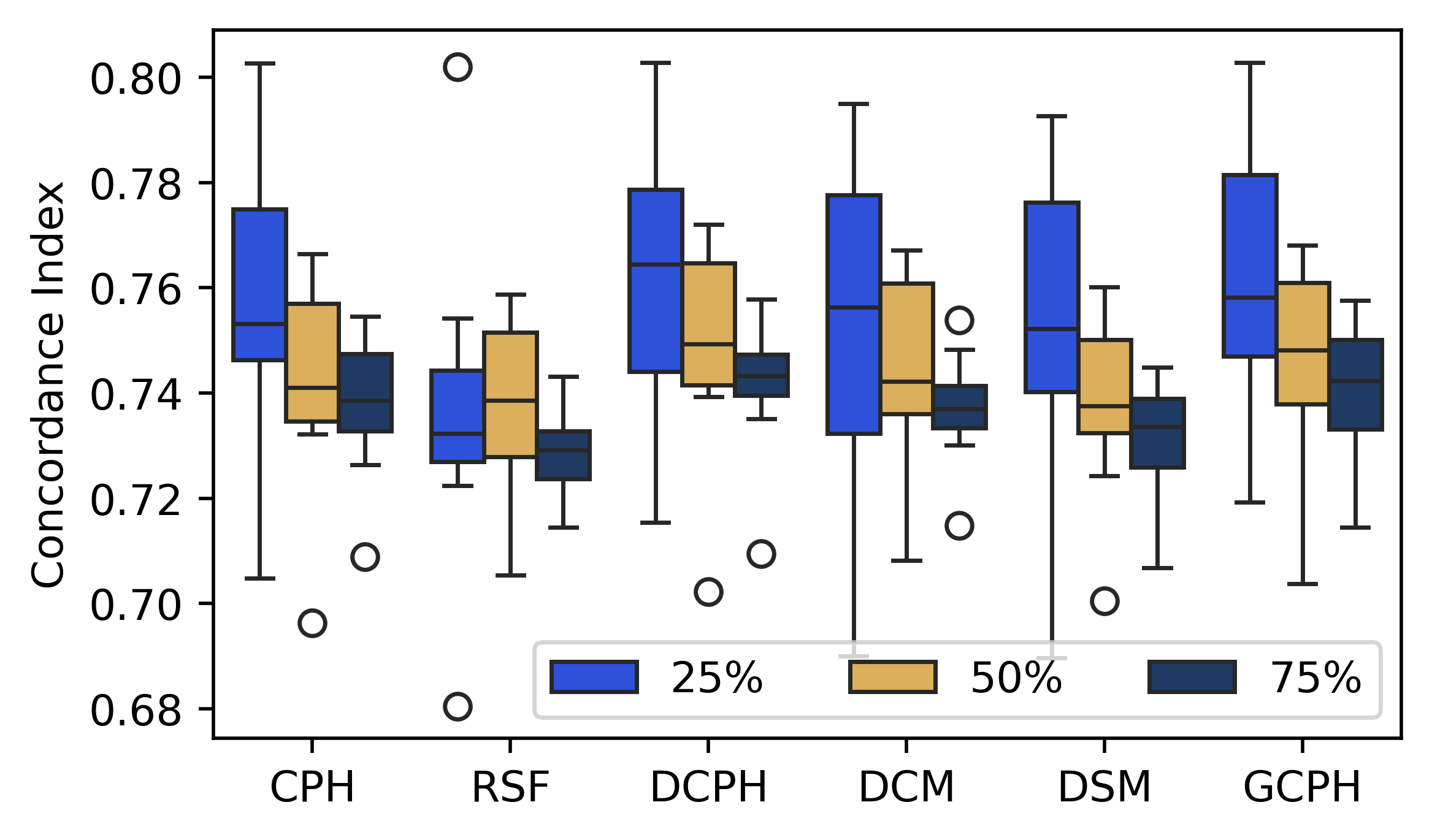}
}%
\hfil
\subfigure[COLON]{
\centering
\includegraphics[width=2.2in]{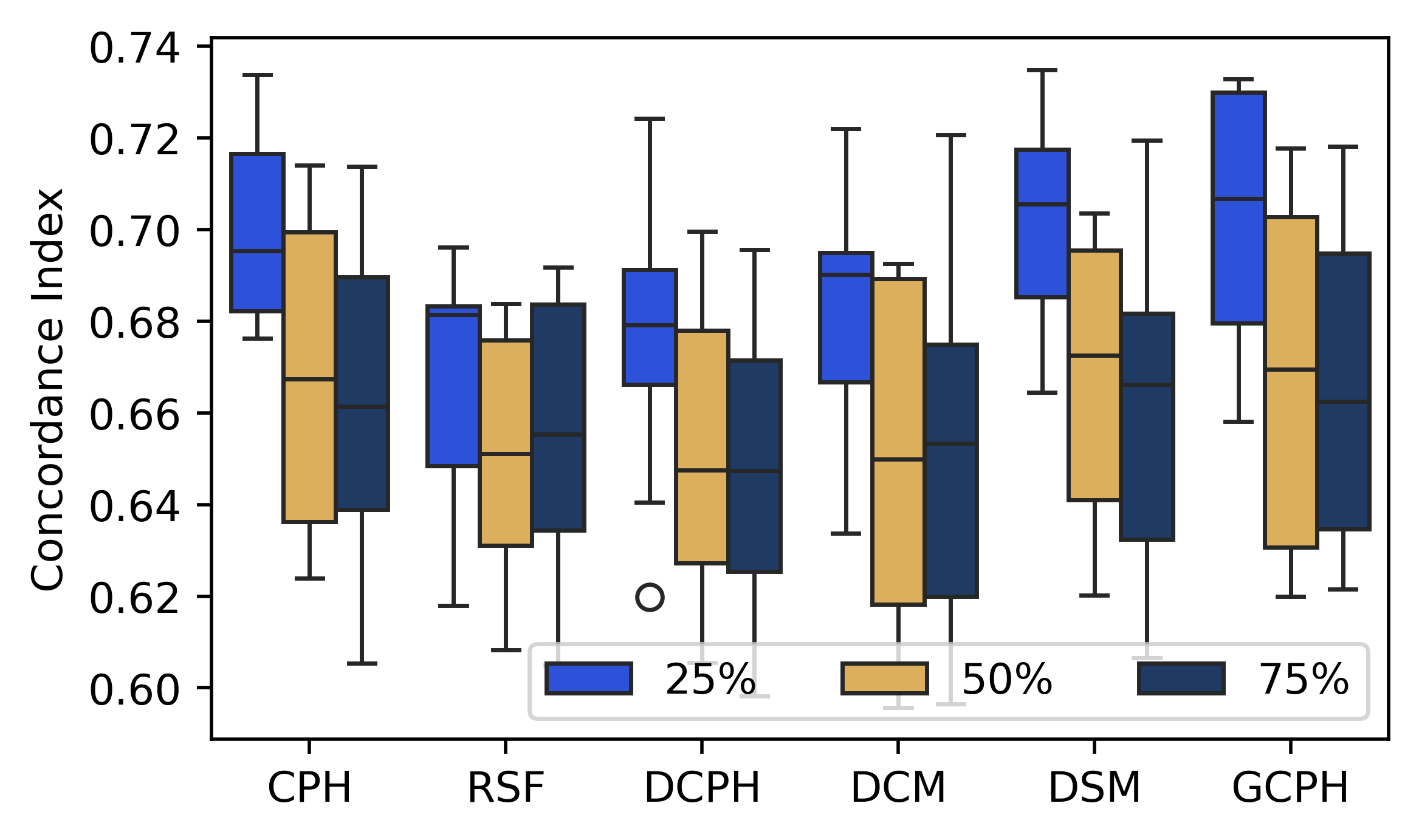}
}%
\hfil
\subfigure[RDATA]{
\centering
\includegraphics[width=2.2in]{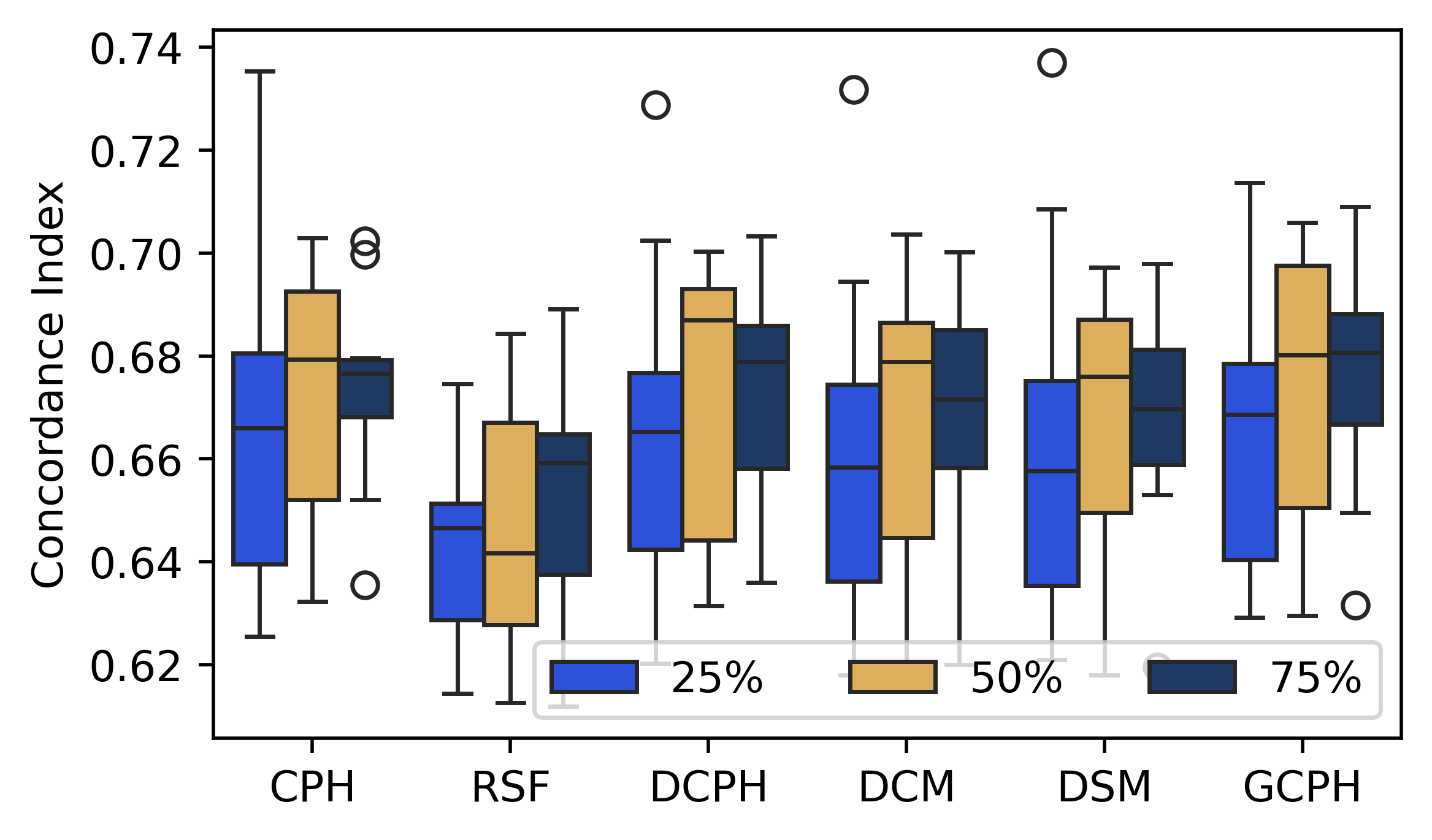}
}%
\hfil
\subfigure[FRTCS]{
\centering
\includegraphics[width=2.2in]{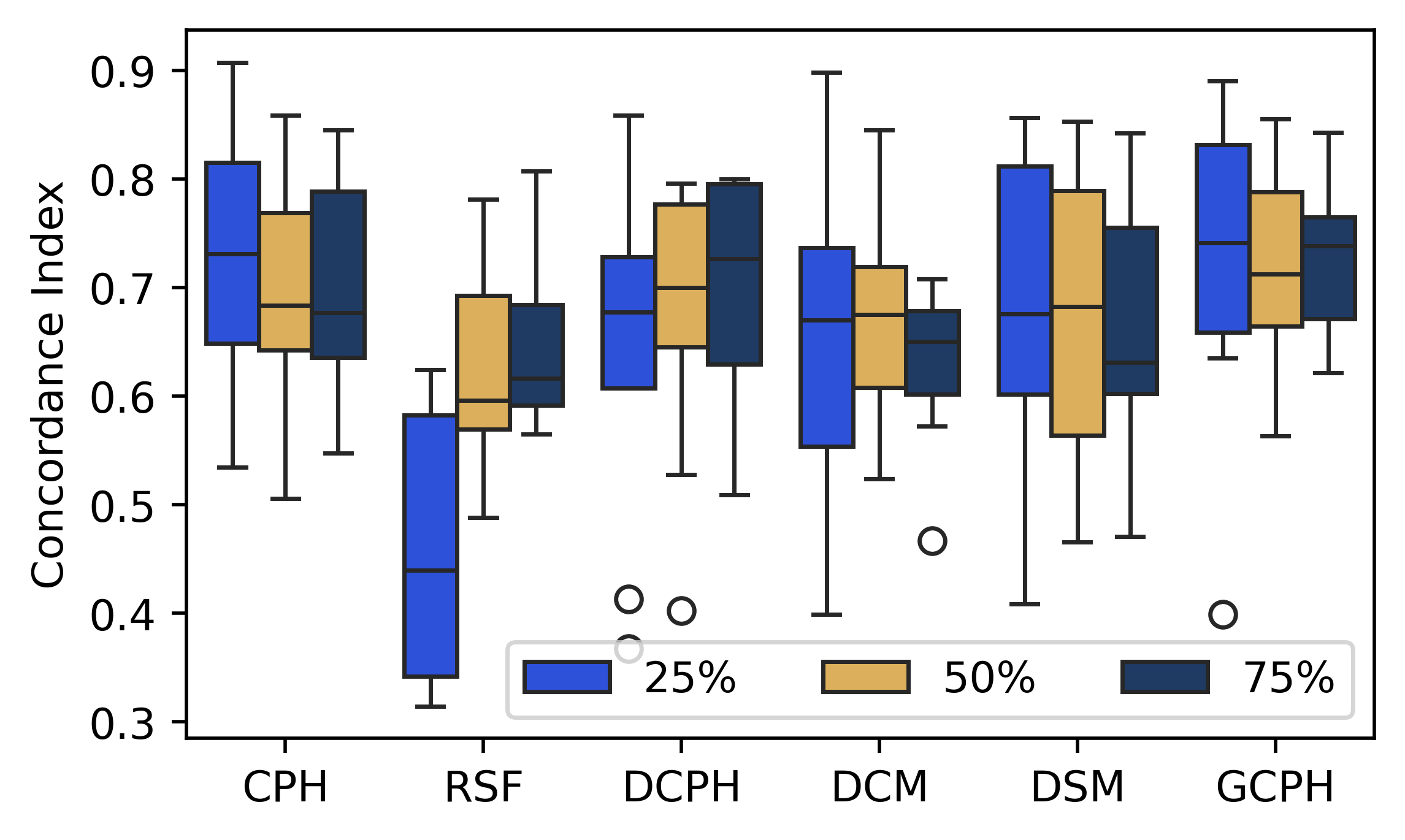}
}%
\caption{\textbf{Box plots of C-Index evaluated on the prediction results by different models}, on various datasets including (a) synthetic linear data, (b) synthetic non-linear data, (c) TRACE, (d) COLON, (e) RDATA, and (f) FRTCS.}
\label{fig:box}
\end{figure*}

\section{Experiments}

We conducted extensive experiments with our proposed model on both synthetic data and public benchmark and compared to different baseline models. 

\subsection{Datasets}

We first generate synthetic data following practices in~\cite{katzman_deepsurv_2018} with known linear and non-linear log-risk.

\subsubsection{Synthetic Linear Data}
We generate synthetic data where subjects have a linear log-risk function given by:
\begin{equation}
f(\mathbf{x}) = x_1 + 2x_2.
\end{equation}

\subsubsection{Synthetic Non-Linear Data}
We also generate synthetic data with a known non-linear log-risk function:
\begin{equation}
f(\mathbf{x}) = \ln(\lambda) \exp\left(-\frac{x_1^2 + x_2^2}{2r^2}\right),
\end{equation}
where we set the parameters $\lambda=5$ and $r=2$, also aligned with the settings in~\cite{katzman_deepsurv_2018}. 

In both synthetic datasets, each $x_v$ is simulated from a \textit{uniform distribution}, $i.e.$, $U(-1,1)$. The initial death time $t_0$ is simulated from an \textit{exponential distribution} with a mean of 5, while the death time $t$ is derived as $t = t_0 / \exp(f(\mathbf{x}))$. The simulated time is then capped so that 10\% of subjects are censored with no events observed~\cite{katzman_deepsurv_2018}.

\subsubsection{TRACE}
The TRACE dataset studies survival probability of patients after myocardial infarction~\cite{jensen_does_1997} and includes 1,878 patients~\cite{martinussen_timereg_2023} with a censoring rate of 48.99\% ($i.e.$, percentage of patients without events occured). It features four binary variables: \textit{sex} (1 if female), clinical heart pump failure (\textit{chf}, 1 if present), \textit{diabetes} (1 if present), and ventricular fibrillation (\textit{vf}, 1 if present). Additionally, it contains two numerical variables: \textit{wmi} (a measure of heart pumping effect based on ultrasound, where 2 is normal and 0 is worst) and \textit{age}.

\subsubsection{COLON}
The COLON dataset examines adjuvant chemotherapy for colon cancer~\cite{moertel_fluorouracil_1995,therneau_survival_2024} and comprises 929 patients with a censoring rate of 51.35\%. It includes five binary variables: \textit{sex} (M if male, F if female), obstruction of colon by tumor (\textit{obstruct}, Y or N), perforation of colon (\textit{perfor}, Y or N), adherence to nearby organs (\textit{adhere}, Y or N), and more than 4 positive lymph nodes (\textit{node4}, Y or N). There are also two categorical variables: treatment (\textit{rx}, with \textit{Obs} for observation, \textit{Lev} for Levamisole, and \textit{Lev+5-FU} for Levamisole+5-FU) and tumor differentiation (\textit{differ}, with levels \textit{well}, \textit{moderate}, and \textit{poor}). Additionally, it contains two numerical variables: \textit{age} and number of lymph nodes with detectable cancer (\textit{nodes}).

\subsubsection{RDATA}
The RDATA dataset includes 1,040 subjects with a censoring rate of 47.40\%~\cite{perme_relsurv_2022}. It features one categorical variable: \textit{agegr} (age group), one binary variable: \textit{sex} (1 if male, 2 if female), and two numerical variables: \textit{age} and date of diagnosis (\textit{year}).

\subsubsection{FRTCS}
The French Three Cities Study (FRTCS) dataset contains 697 subjects with a censoring rate of 89.67\%~\cite{hosmer_applied_2008}. It includes four binary variables: \textit{sex} (M if male, F if female) and use of antihypertensive drugs (\textit{antihyp0} to \textit{antihyp2}, Y or N). Additionally, it features nine numerical variables: \textit{age}, three records of systolic blood pressure (\textit{sbp0} to \textit{sbp2}), three records of diastolic blood pressure (\textit{dbp0} to \textit{dbp2}), and two records of dates (\textit{date0} and \textit{date1}).

\subsection{Model Settings}
In our model, the number of activation functions is set to match the number of covariates in each dataset, $i.e.$, $V$. The weights for the $L1$ norm and entropy regularization losses are set to $\mu_1 = 1$ and $\mu_2 = 10$ in Equation \ref{eq:reg}. Additionally, the weight for the total regularization loss is set to $\gamma = 0.1$ in Equation \ref{eq:loss}. And for spline functions, we set an order of 3, $i.e.$, $k=3$ in Equation \ref{eq:spline}, and a total of 5 intervals for each spline function. We select the symbolic functions as $y(x)$ for approximation, including $x$, $x^2$, $x^3$, $x^4$, $exp$, $ln$, $sqrt$, $tanh$, $sin$. We also test \textit{Linear GCPH} by using only $x$ to fit activation functions, denoted as GCPH-$l$.

The baseline models we experimented and compared with include CPH~\cite{cox_regression_1972}, RSF~\cite{ishwaran_random_2008}, DCPH~\cite{katzman_deepsurv_2018}, DCM~\cite{nagpal_deep_2021-1}, and DSM~\cite{nagpal_deep_2021}.

\subsection{Evaluation Metrics}

\subsubsection{Concordance Index (C-Index)}
The C-Index assesses how well the predicted risk scores align with the actual survival times.
Given a dataset with $n$ pairs of instances $(i, j)$, where each instance $i$ has a survival time $t_i$ and an event indicator $\delta_i$, and $\hat{r}_i$ denotes the predicted risk score for instance $i$, the C-Index is defined as:
\begin{equation}
\text{CI} = \frac{\sum_{i < j} I(t_i < t_j) \cdot I(\hat{r}_i > \hat{r}_j) \cdot (\delta_i + \delta_j)}{\sum_{i < j} I(t_i < t_j) \cdot (\delta_i + \delta_j)},
\end{equation}
where $I(\cdot)$ is an indicator function that returns 1 if the condition is met and 0 otherwise. The numerator counts the number of concordant pairs, and the denominator normalizes by the total number of comparable pairs.

\subsubsection{Brier Score}
It measures the accuracy of probabilistic predictions, taking into account both the calibration and discrimination of the model. It is defined as the \textit{mean squared error} between the predicted survival probability and the actual outcome.
For a survival model, the Brier Score at a specific time point $t$ can be written as:
\begin{equation}
\text{Brier Score}(t) = \frac{1}{n} \sum_{i=1}^n \frac{\left( \hat{S}(t \,|\, \mathbf{x}_i) - I(t_i > t) \right)^2}{\hat{G}(t \,|\, \mathbf{x}_i)},
\end{equation}
where $\hat{S}(t \,|\, \mathbf{x}_i)$ is the predicted survival probability for instance $i$ at time $t$.
And $\hat{G}(t \,|\, \mathbf{x}_i)$ is the Kaplan-Meier estimate of the survival function of the censoring distribution. 

In our experiments, we choose the $25$, $50$, and $75$-\textit{th} percentiles of time $t_i$ in the train data as the time $t$ to compare the Brier Score, as well as the C-Index.
A higher C-Index indicates better performance of the model and a C-Index of 0.5 corresponds to random chance. A lower Brier Score indicates better accuracy of the probabilistic predictions.

\begin{figure}[!t]\centering
\subfigure[Ground Truth]{
\label{fig:linear-true}
\centering
\includegraphics[width=1.2in]{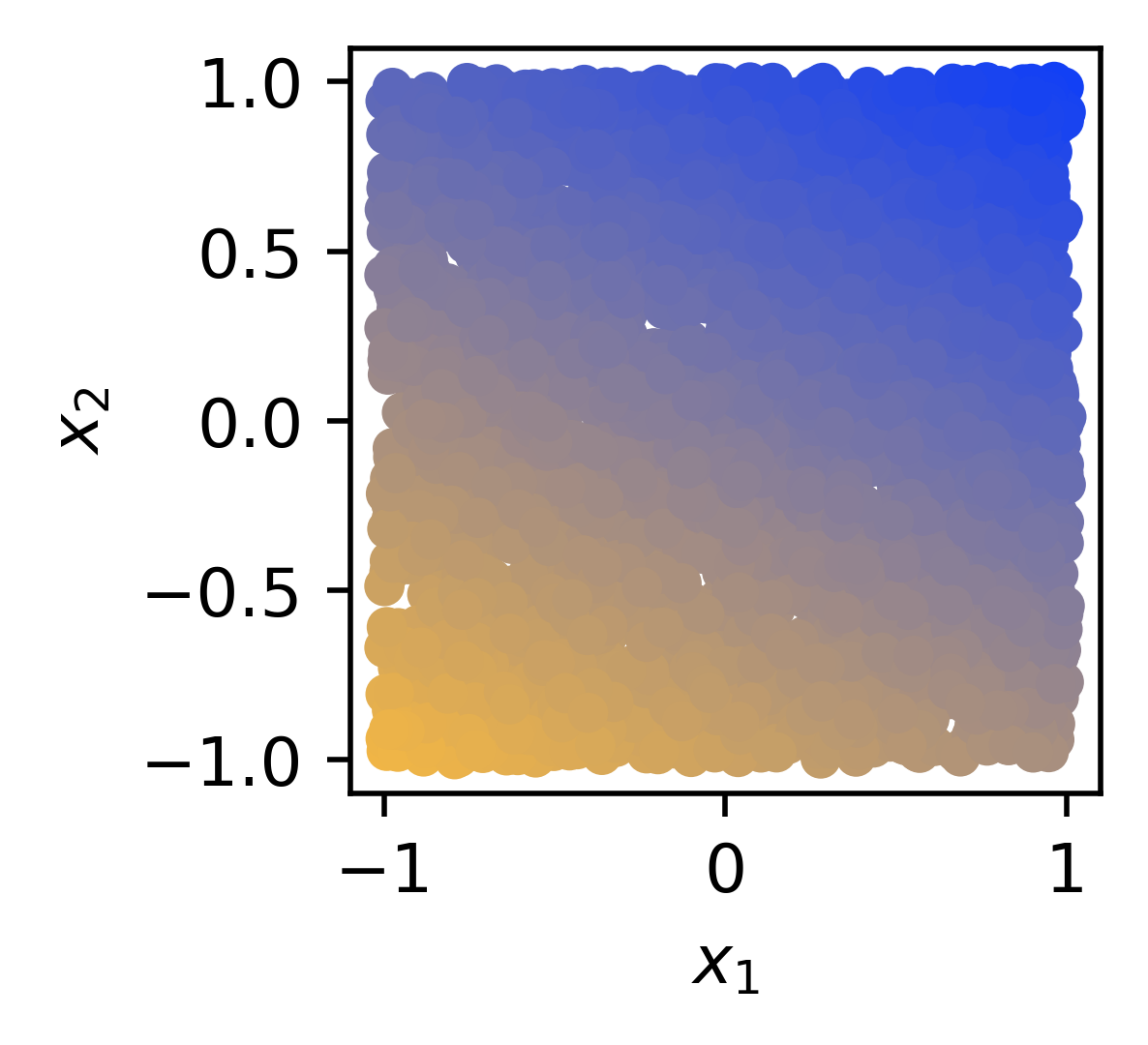}
}%
\hfil
\subfigure[CPH]{
\centering
\includegraphics[width=1.2in]{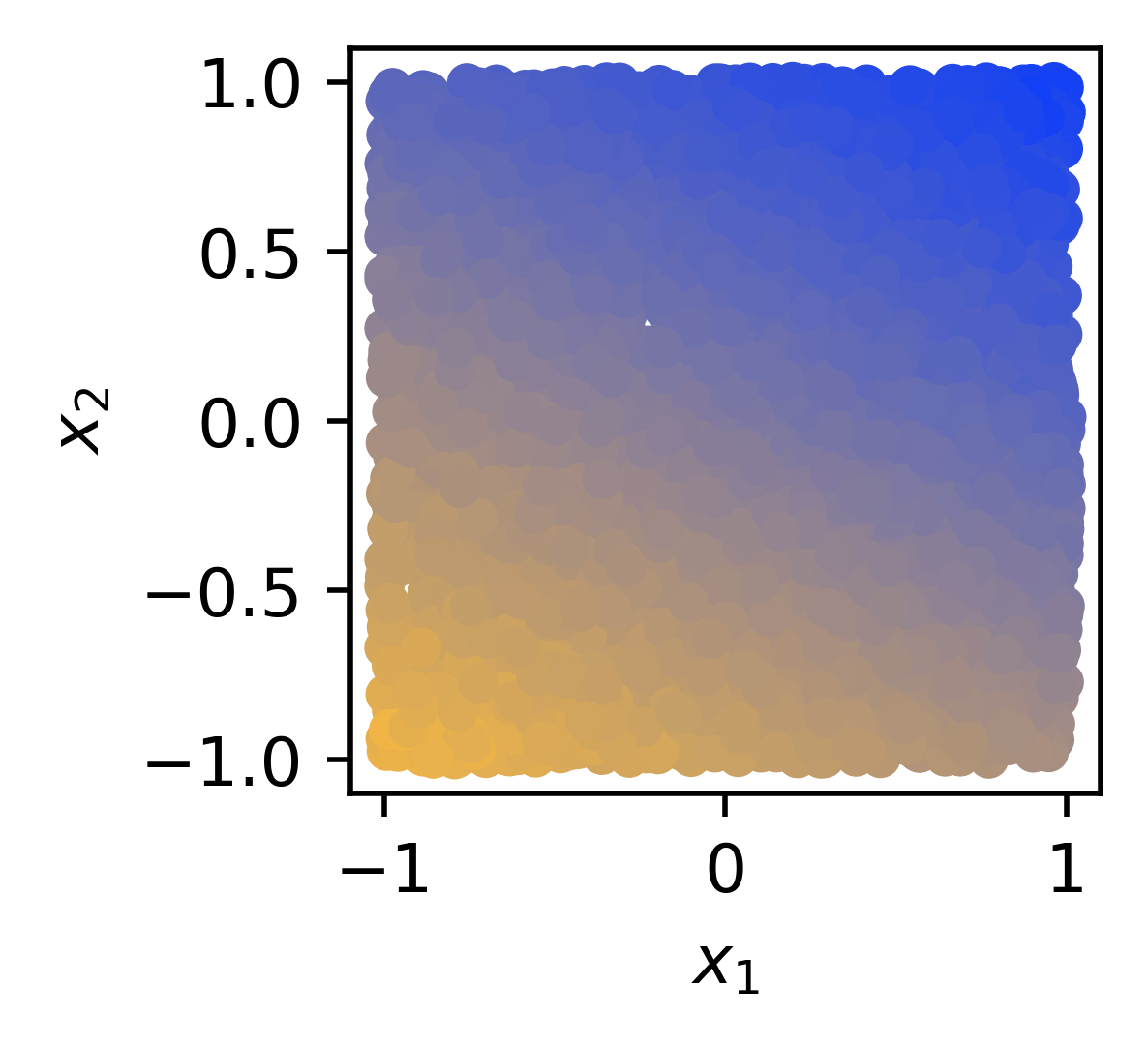}
}%
\hfil

\subfigure[DCPH]{
\label{fig:linear-dcph}
\centering
\includegraphics[width=1.2in]{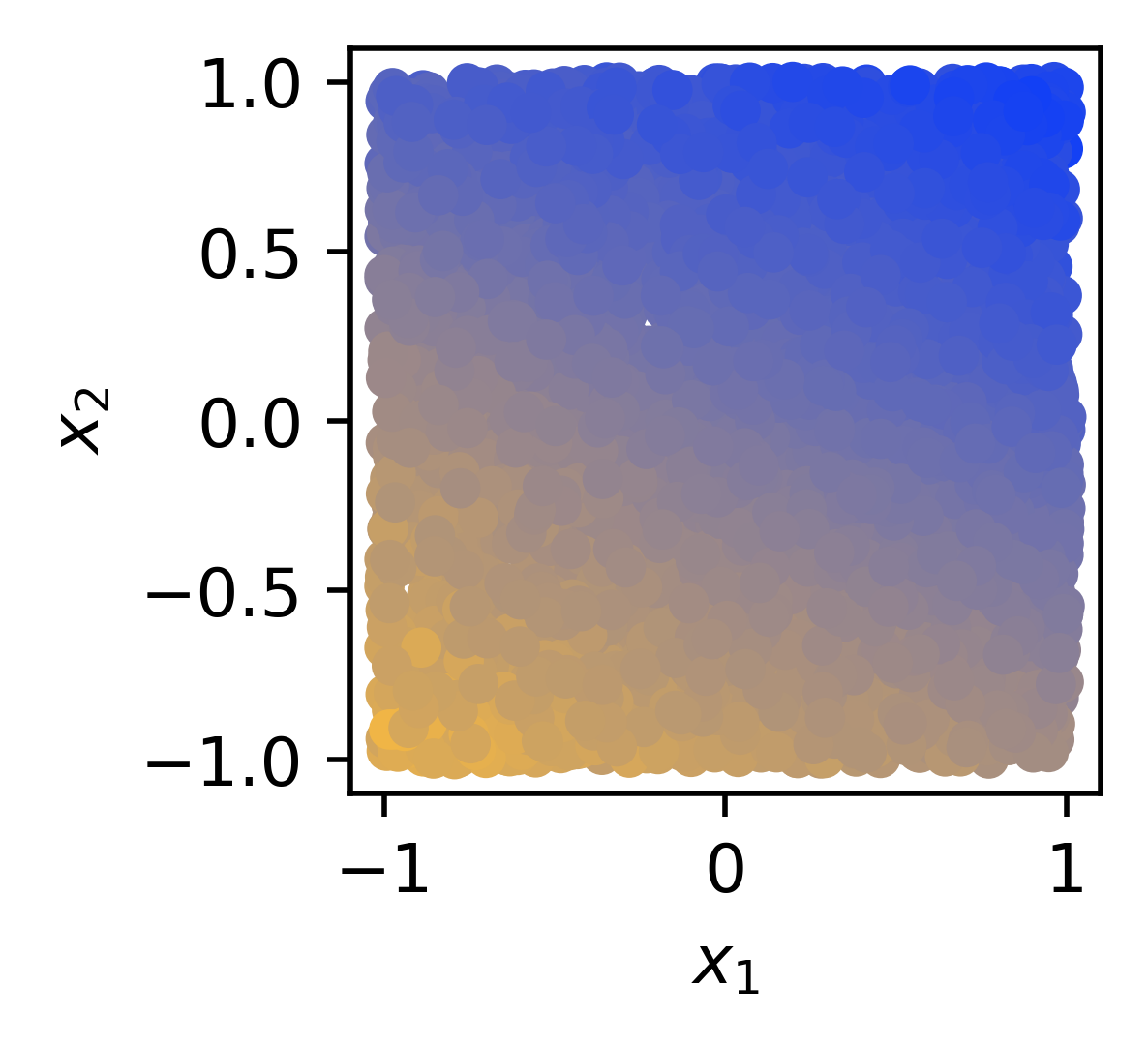}
}%
\hfil
\subfigure[GCPH]{
\label{fig:linear-coxkan}
\centering
\includegraphics[width=1.2in]{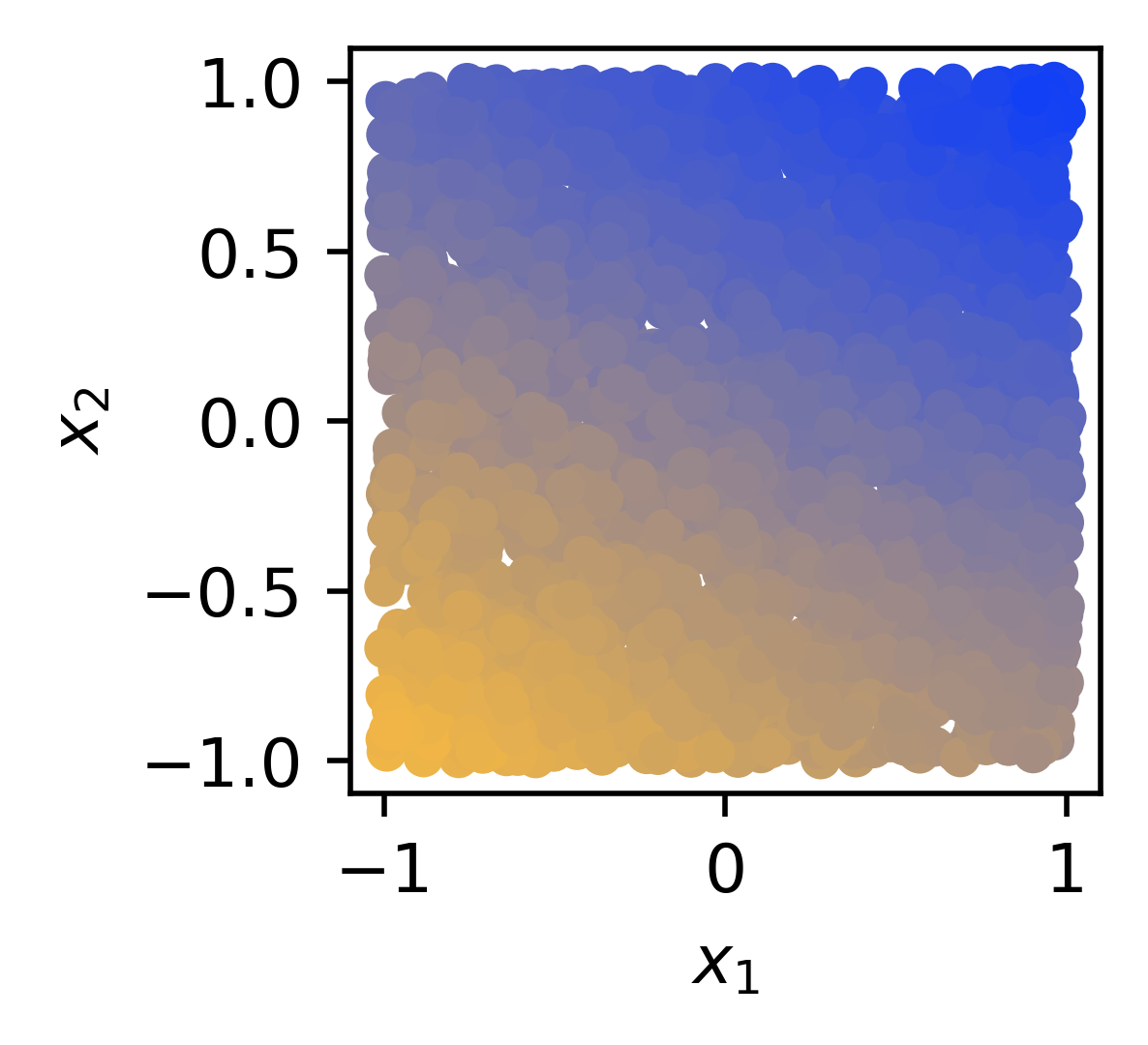}
}%
\caption{\textbf{Comparison of learning abilities from linear relationships}. The experiments are based on synthetic non-linear data, illustrated in (a). The models compared are (b) traditional CPH~\cite{cox_regression_1972}, (c) DCPH~\cite{katzman_deepsurv_2018}, and (d) our proposed GCPH model.}
\label{fig:linear}
\end{figure}

\subsection{Experimental Results}

The experimental results using different models on various datasets are summarized in Table \ref{tab:summary} with corresponding C-index and Brier Score evaluated. We also illustrate the results on C-index in Figure \ref{fig:box} with box plots. It can be seen that our proposed model achieves competitive performance.

\subsubsection{Linear and Non-Linear Experiments}
In linear experiments, the CPH model performs well due to its inherent linear assumption, while GCPH-$l$ achieves the optimal performance as shown in Table \ref{tab:summary}. For non-linear experiments, DCPH achieves the best performance, with the proposed GCPH coming in a close second. Despite having much fewer trained parameters (2 hidden layers with 100 neurons adopted for DCPH), GCPH delivers competitive results with its much simpler and more transparent architecture.

\begin{figure}[!t]\centering
\subfigure[$\phi_1$ vs. $\hat{\phi}_1$]{
\centering
\includegraphics[width=1.2in]{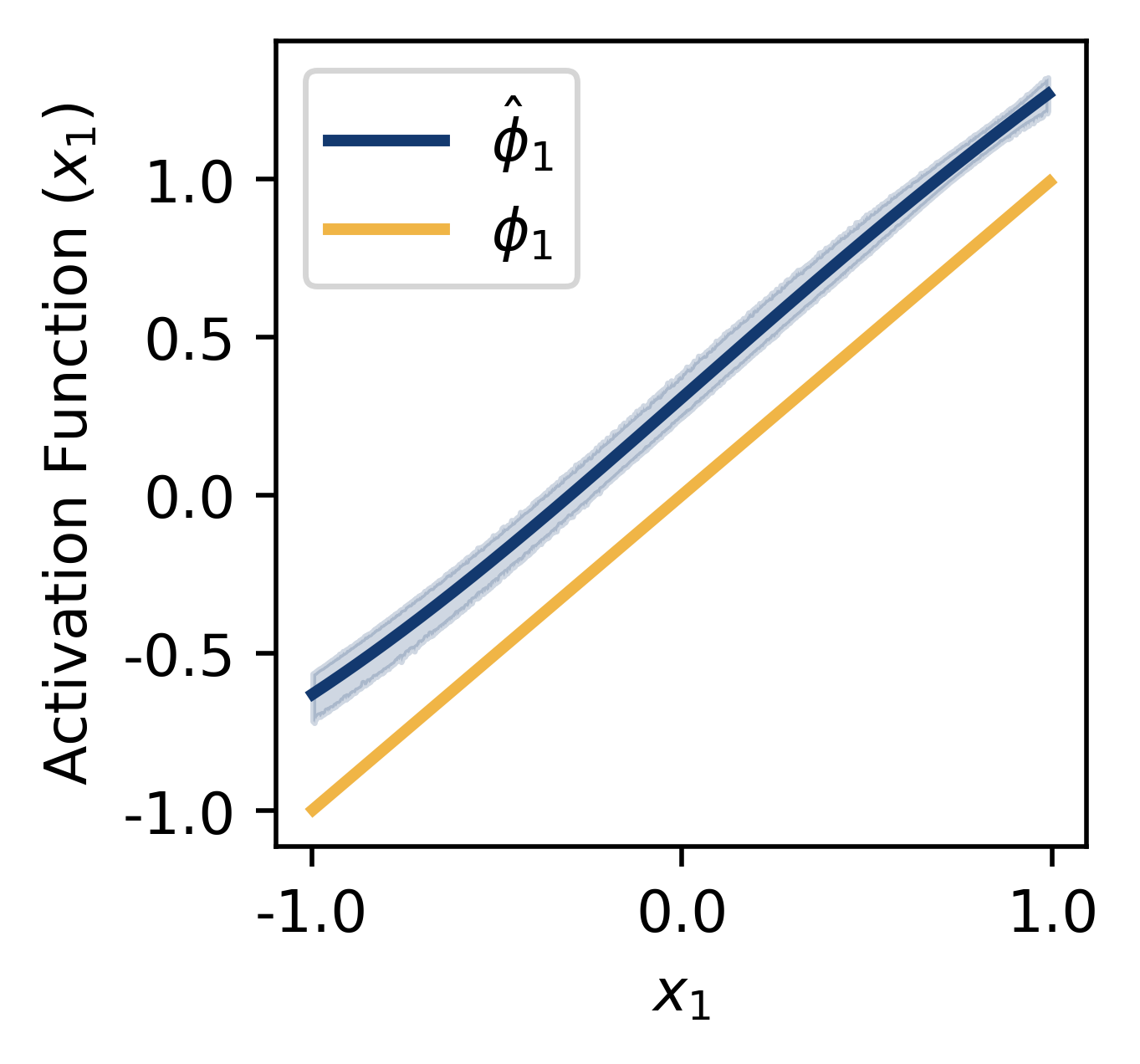}
}%
\hfil
\subfigure[$\phi_2$ vs. $\hat{\phi}_2$]{
\centering
\includegraphics[width=1.2in]{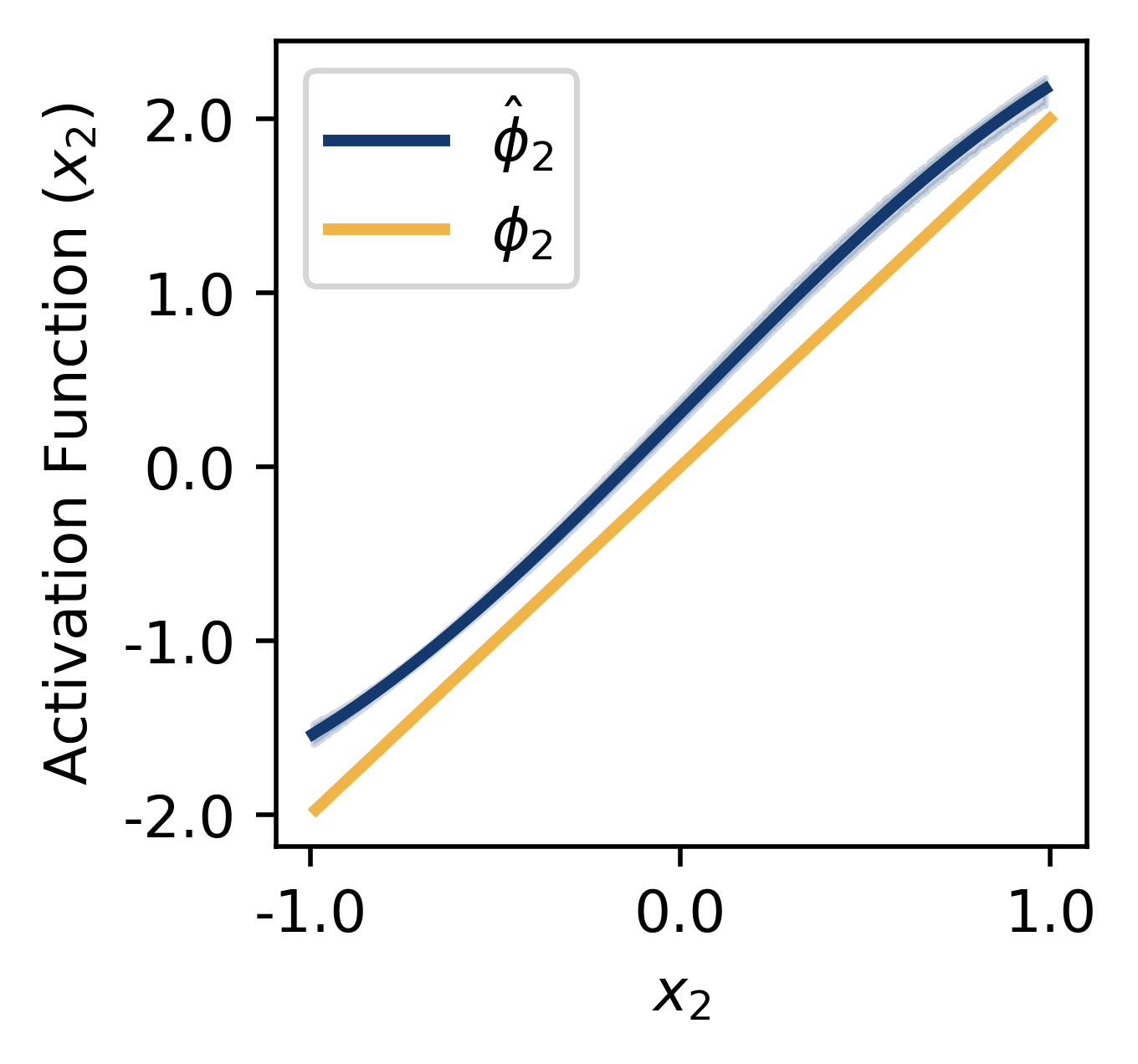}
}%
\hfil
\subfigure[$\phi_1$ vs. $\hat{\phi}_1$]{
\centering
\includegraphics[width=1.2in]{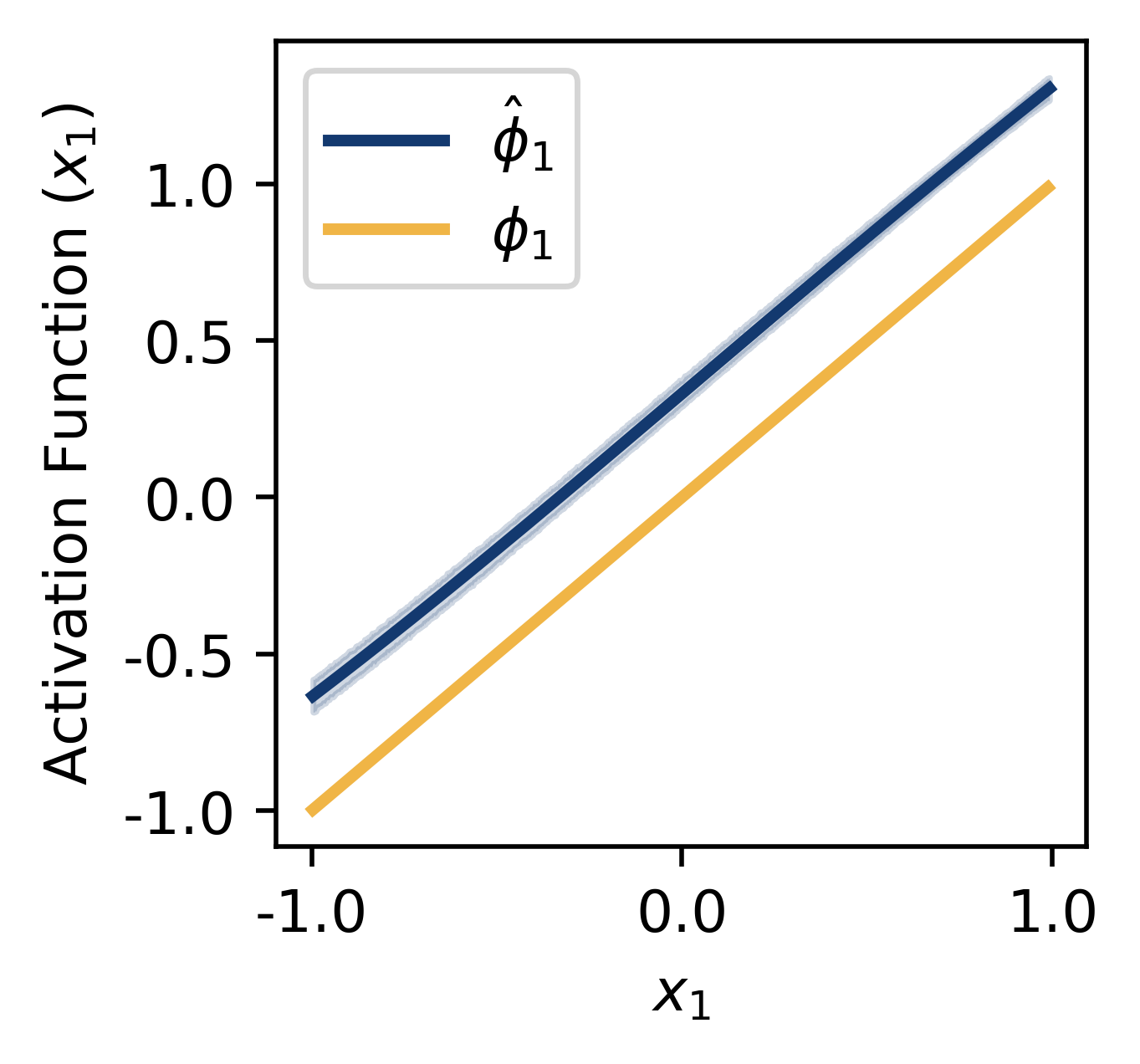}
}%
\hfil
\subfigure[$\phi_2$ vs. $\hat{\phi}_2$]{
\centering
\includegraphics[width=1.2in]{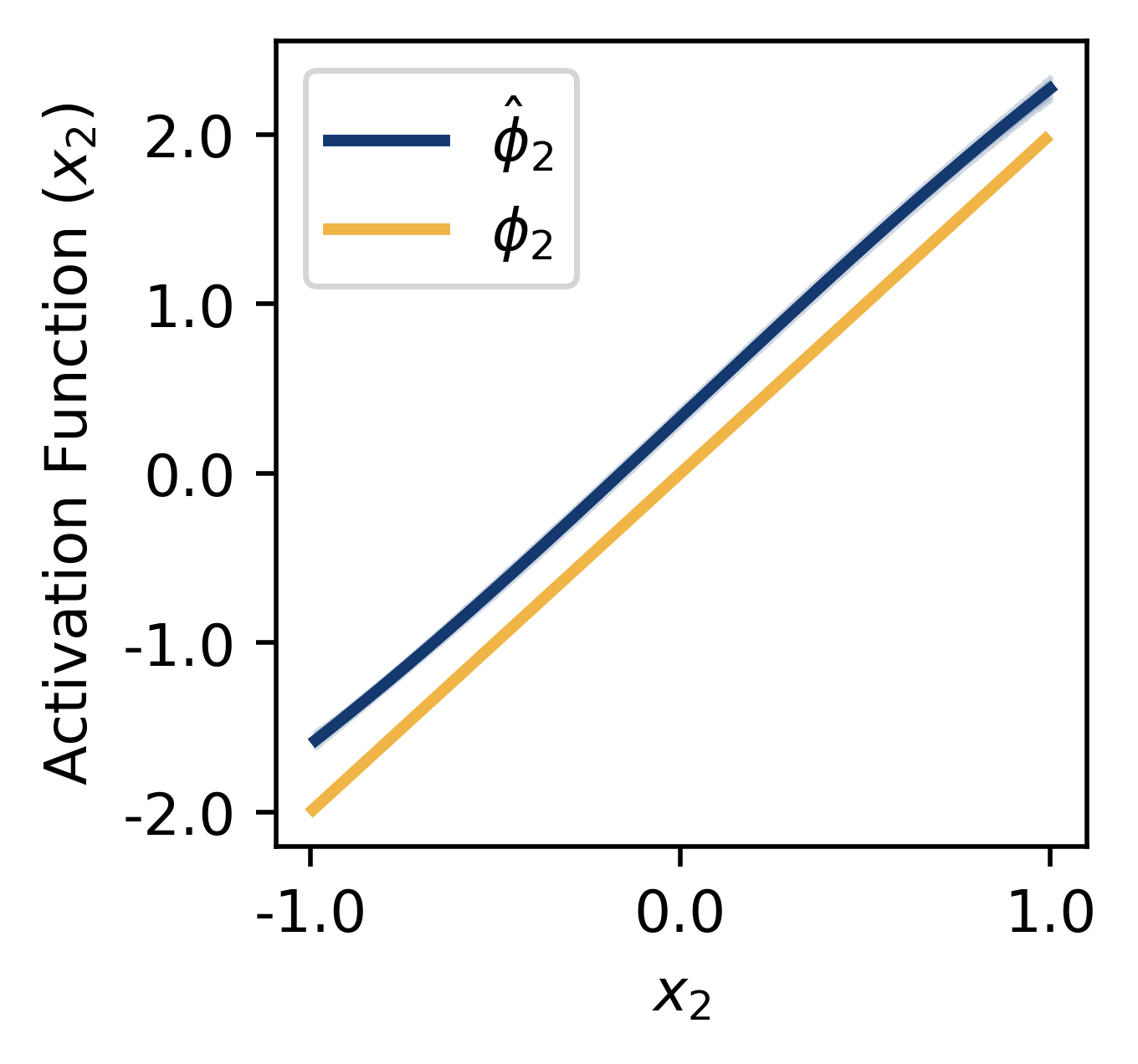}
}%
\caption{\textbf{Symbolic functions learned by GCPH ($a$-$b$) and GCPH-$l$ ($c$-$d$) compared to ground truth in the linear experiments}. $\phi_1(x)=x$ and $\phi_2(x)=2x$, while $\hat{\phi}_1$ and $\hat{\phi}_2$ are the predicted ones with multiple runs.}
\label{fig:linearf}
\end{figure}

\begin{figure}[!t]\centering
\subfigure[$\phi_1$ vs. $\hat{\phi}_1$]{
\centering
\includegraphics[width=1.2in]{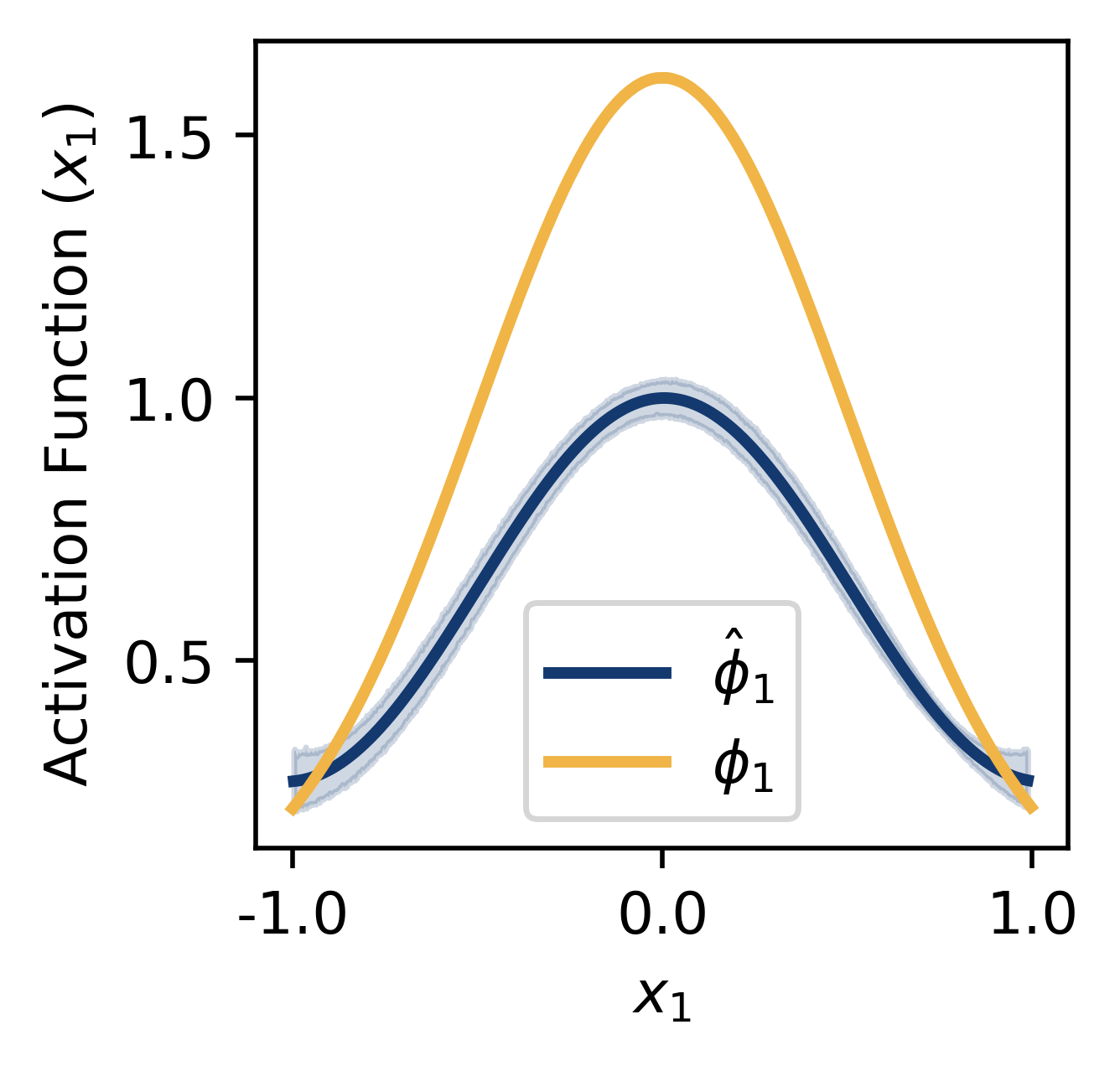}
}%
\hfil
\subfigure[$\phi_2$ vs. $\hat{\phi}_2$]{
\centering
\includegraphics[width=1.2in]{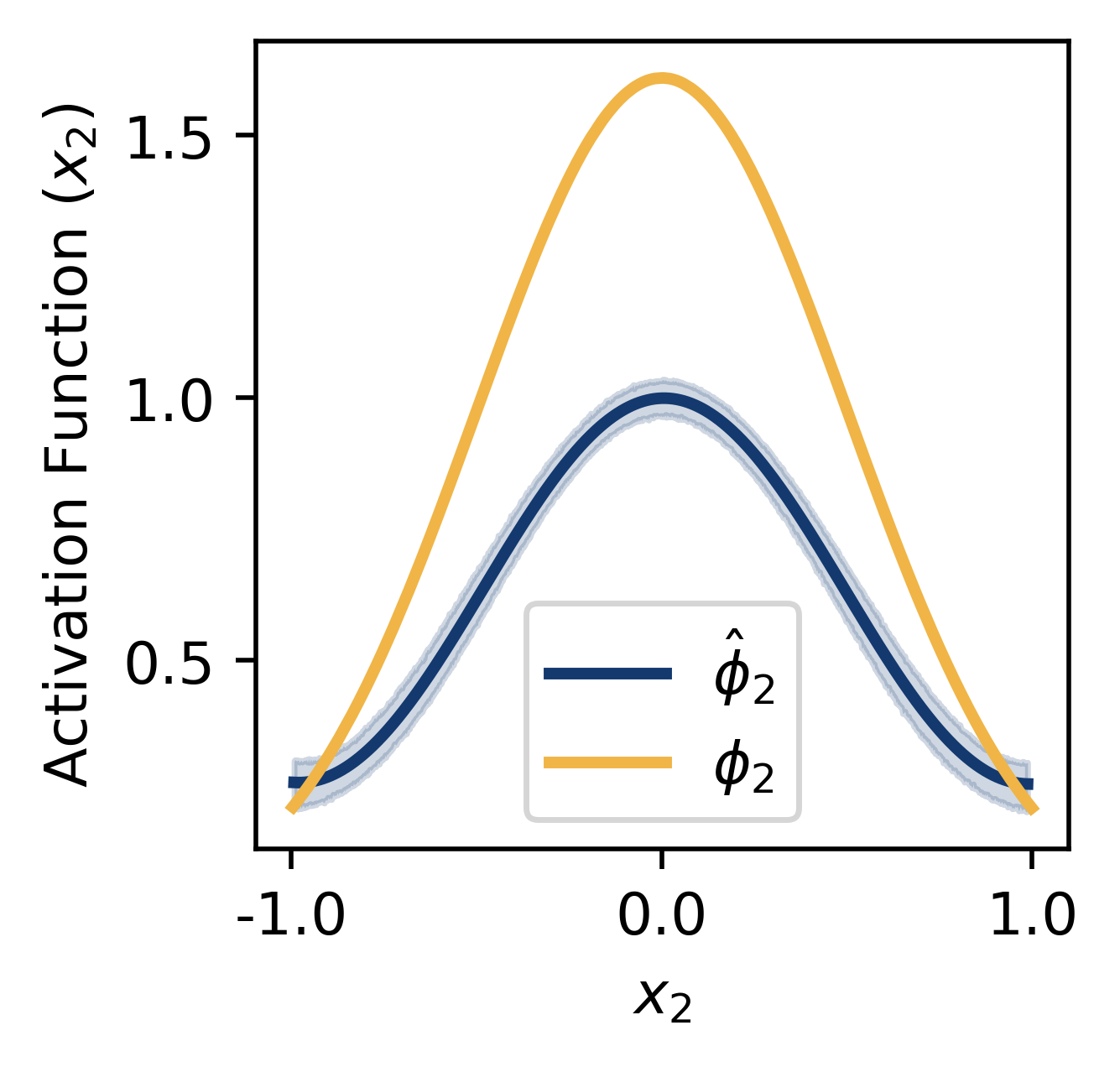}
}%
\caption{\textbf{Symbolic functions learned by GCPH compared to ground truth in the non-linear experiments}. $\phi_1(x)=\phi_2(x)=\ln(\lambda) \exp{(-0.5x^2 / r^2)}$, while $\hat{\phi}_1$ and $\hat{\phi}_2$ are the predicted ones with multiple runs.}
\label{fig:nonlinearf}
\end{figure}

\begin{figure}[!t]\centering
\subfigure[$\hat{\phi}_1(x_1)$]{
\centering
\includegraphics[width=1in]{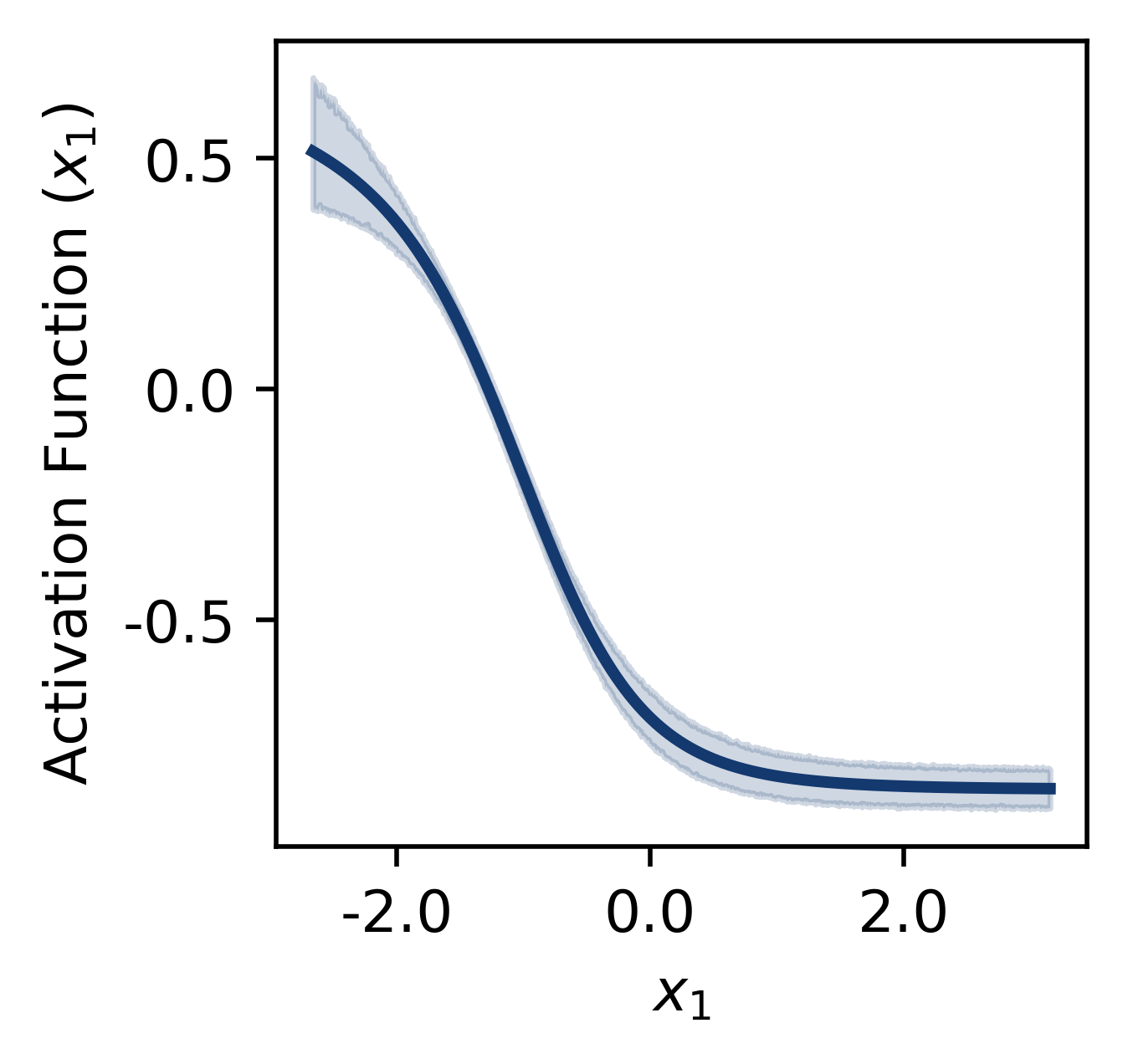}
}%
\hfil
\subfigure[$\hat{\phi}_2(x_2)$]{
\centering
\includegraphics[width=1in]{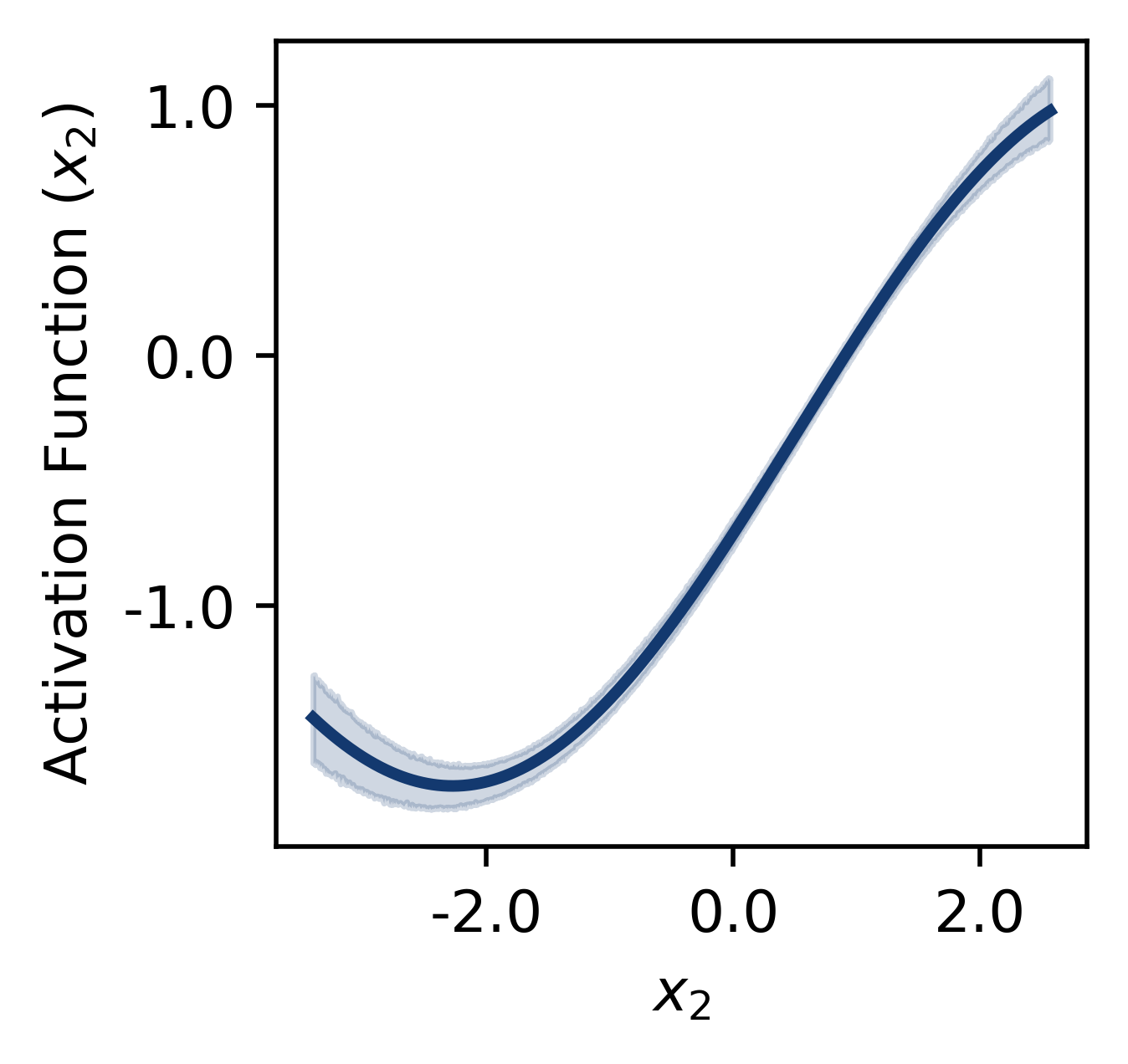}
}%
\hfil
\subfigure[$\hat{\phi}_3(x_3)$]{
\label{fig:trace_sym_3}
\centering
\includegraphics[width=1in]{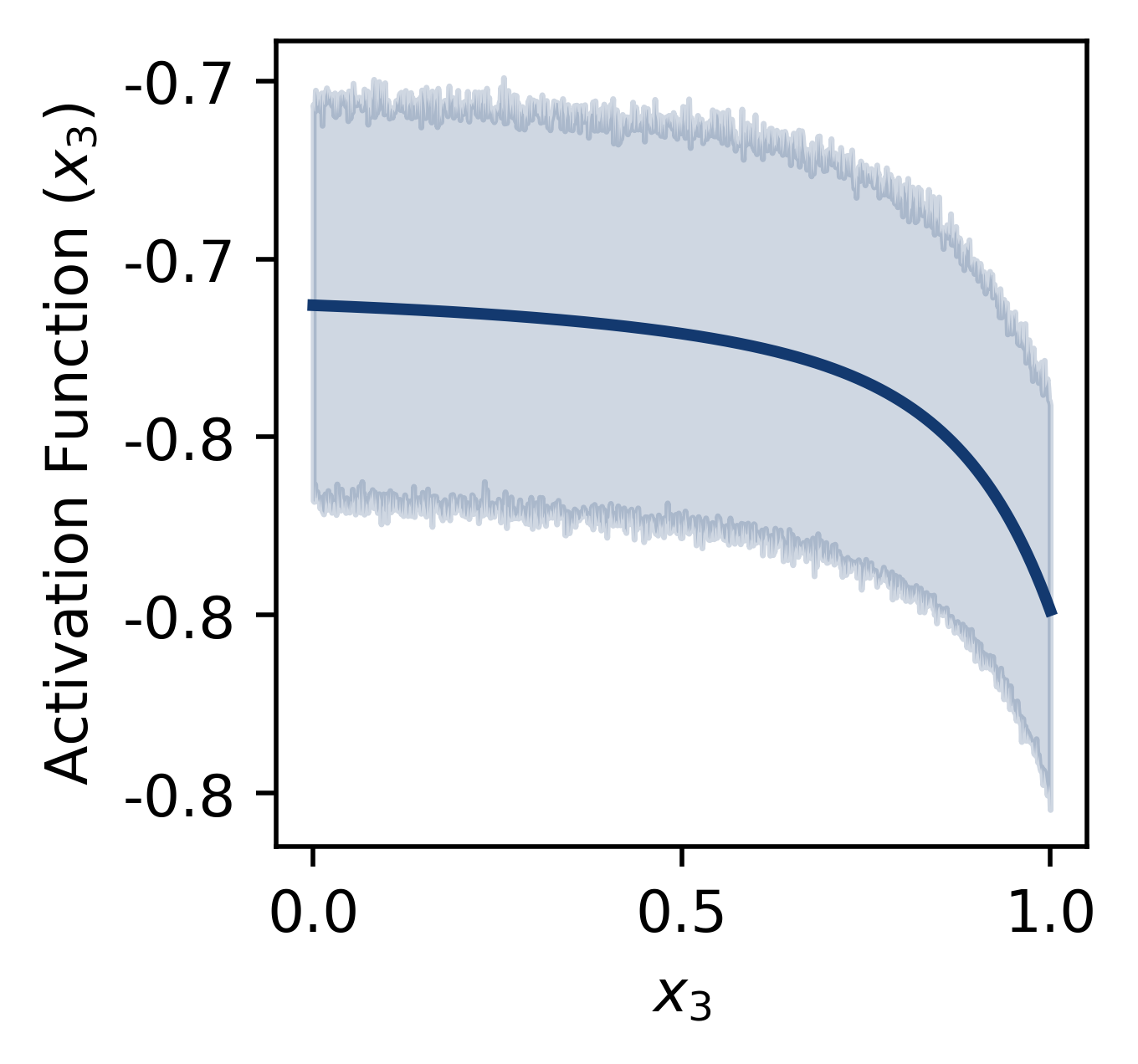}
}%

\subfigure[$\hat{\phi}_4(x_4)$]{
\centering
\includegraphics[width=1in]{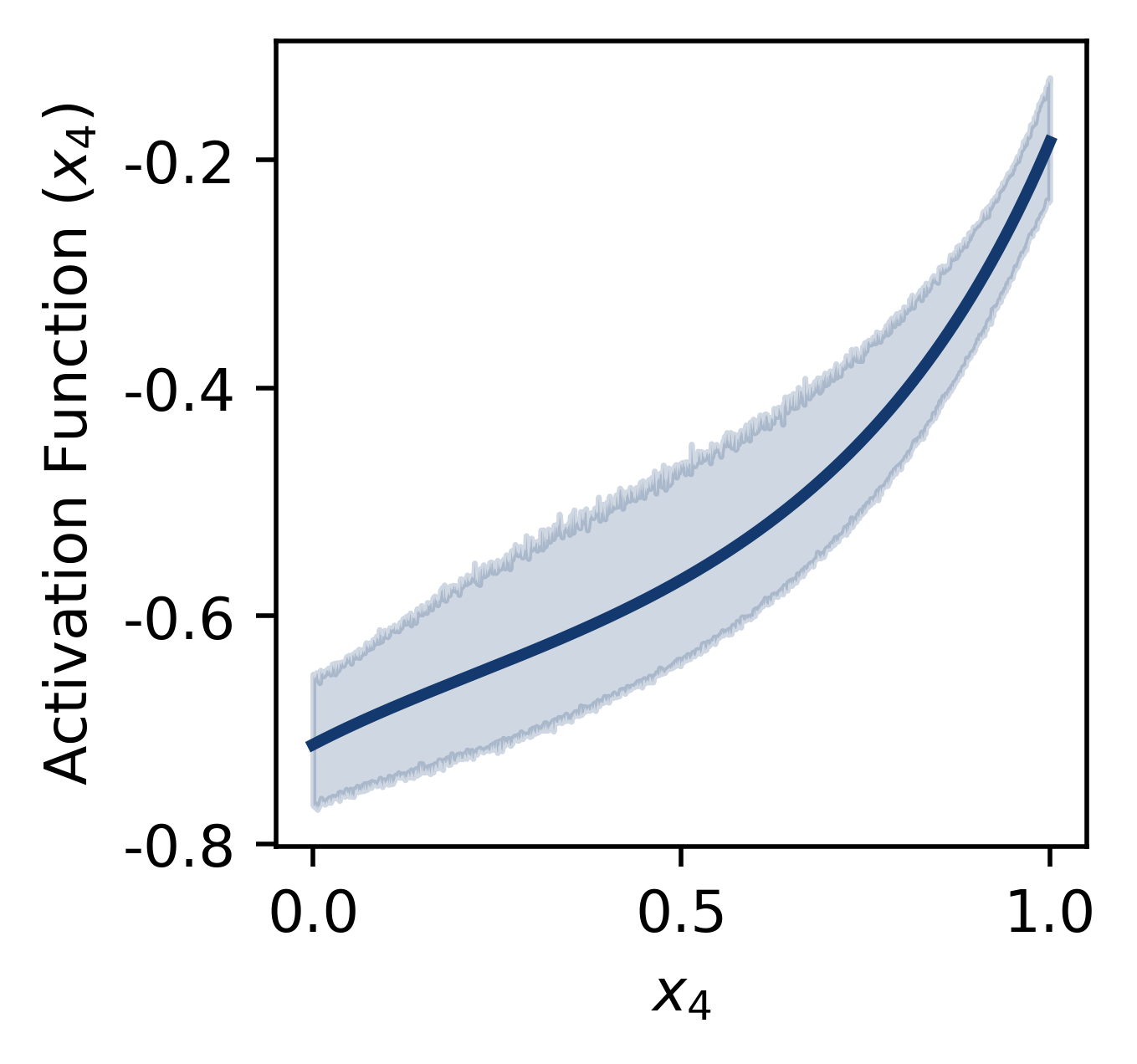}
}%
\hfil
\subfigure[$\hat{\phi}_5(x_5)$]{
\centering
\includegraphics[width=1in]{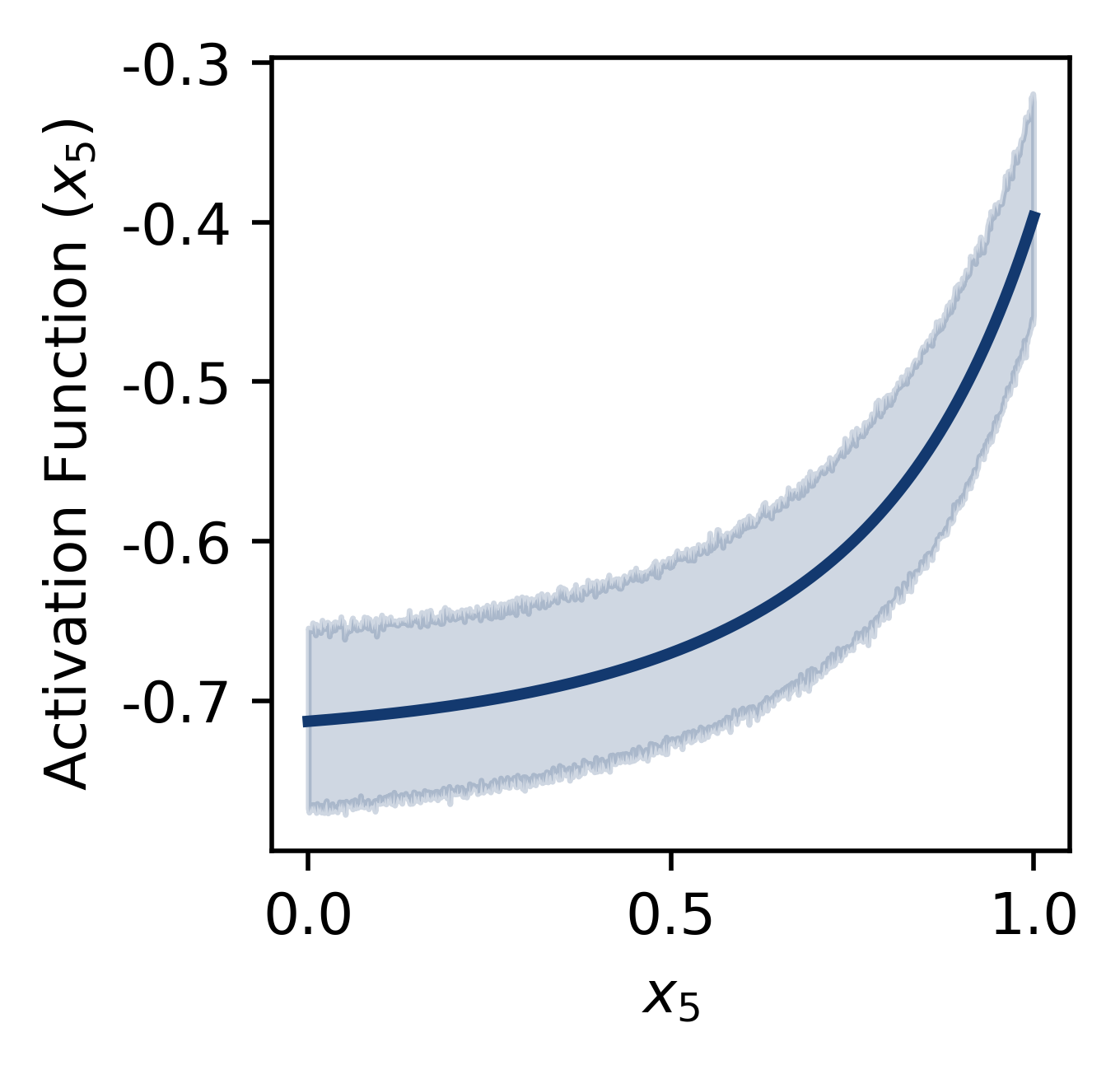}
}%
\hfil
\subfigure[$\hat{\phi}_6(x_6)$]{
\centering
\includegraphics[width=1in]{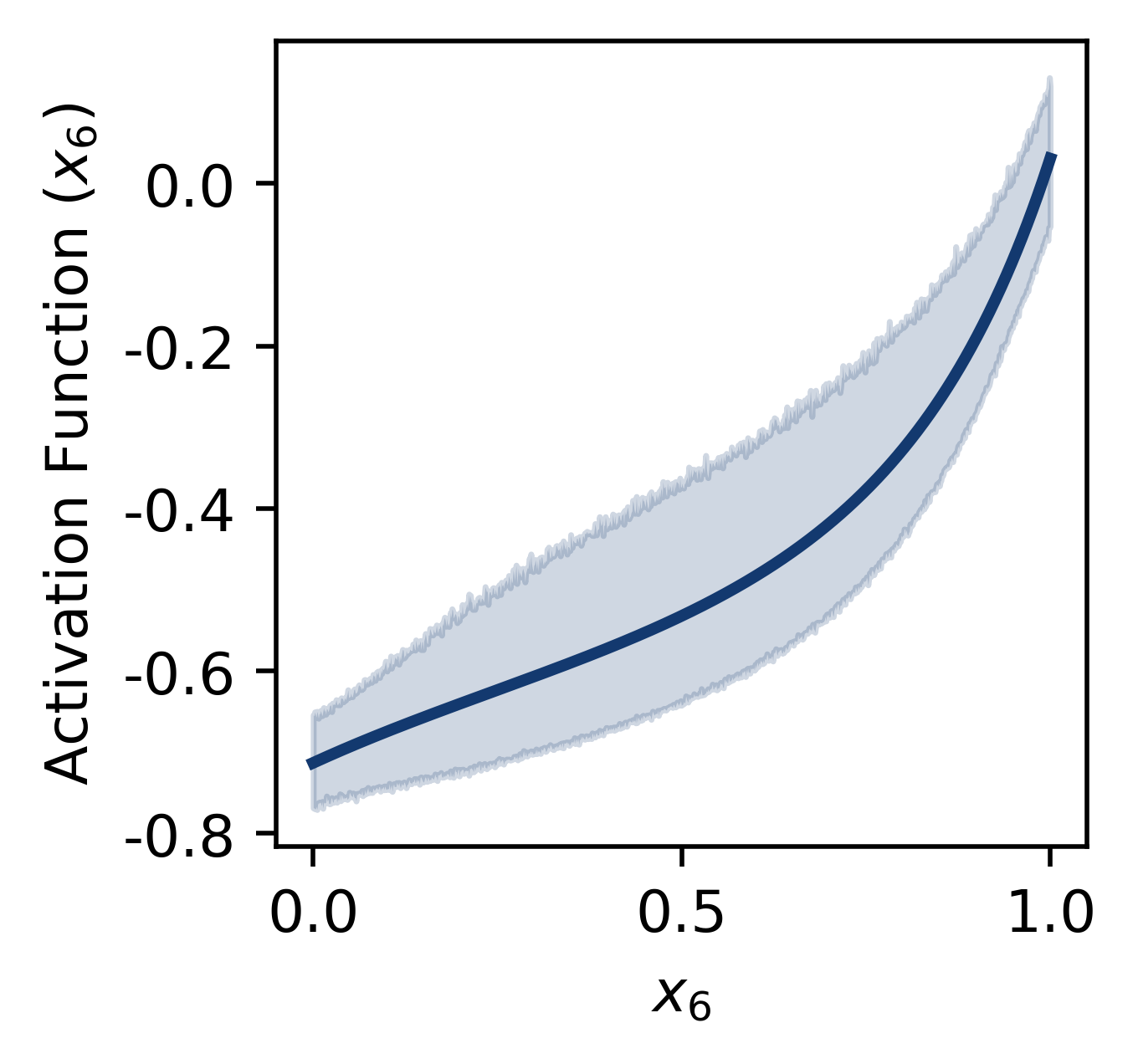}
}%
\caption{\textbf{Symbolic functions learned by GCPH with TRACE dataset}. Variables from (a) $x_1$ to (f) $x_6$ represent \textit{wmi}, \textit{age}, \textit{sex}, presence of \textit{chf}, \textit{diabetes}, and \textit{vf}, respectively.}
\label{fig:trace_sym}
\end{figure}

Figures \ref{fig:linear} and \ref{fig:nonlinear} illustrate the prediction results for linear and non-linear experiments, respectively. In Figure \ref{fig:linear-true}, the true linear relationship is shown, and CPH, DCPH, and GCPH effectively capture this relationship. GCPH's predictions are notably less noisy, as compared between Figures \ref{fig:linear-dcph} and \ref{fig:linear-coxkan}.
In non-linear experiments, depicted in Figure \ref{fig:nonlinear}, the differences in model performance are more pronounced. The CPH model struggles with the non-linear relationships due to its linear assumptions, while GCPH closely matches the ground truth shown in Figure \ref{fig:non-linear-true}.

Figures \ref{fig:linearf} and \ref{fig:nonlinearf} present the symbolic functions estimated by GCPH for each variable. Each symbolic function $\hat{\phi}_v(x_v)$ is obtained by setting all other variables in $f(\mathbf{x}; \hat{\bm{\Phi}})$ to $0$, $i.e.$, $\hat{\phi}_v(x_v) = f(x_v, x_{s:s \neq v} = 0; \hat{\bm{\Phi}})$. These symbolic functions closely approximate the ground truth, effectively reflecting the actual shape of the relationships.

\begin{table*}[!t]
\centering\fontsize{9}{11}\selectfont
\caption{\textbf{Summary of symbolic non-linear log-risk function approximated with different public benchmark}. Two floating digits are reserved to simplify the symbolic functions while showing the most important features.}
\begin{tabular}{>{\centering}p{1.5cm}l}
\hline
Data & \textbf{\textit{Symbolic Non-Linear Log-Risk Function}}\\
\hline

TRACE & 

$\begin{aligned}[t]
        f(\mathbf{x}; \hat{\bm{\Phi}}) =- 1.6\text{tanh}(0.5x_1 + 1.1) -1.3\text{sin}(0.5x_2 + 9.3)  + 0.9
      \end{aligned}$, $\quad x_1$: \textit{wmi}, $x_2$: \textit{age} \\
      
\hline
COLON & 

$\begin{aligned}[t]
        f(\mathbf{x}; \hat{\bm{\Phi}}) = 0.2\text{tanh}(1.4x_1 - 0.6) + 0.5\text{tanh}(0.7x_2) - 0.2
      \end{aligned}$, $\quad x_1$: \textit{age}, $x_2$: \textit{number of lymph nodes} \\
\hline
\multirow{ 2}{*}{RDATA} & 

$\begin{aligned}[t]
        f(\mathbf{x}; \hat{\bm{\Phi}}) & =1.6\text{tanh}(0.4x_1 + 0.2) - 0.2\text{tanh}(3.9x_2 - 1.1) + 0.1(x_5 + 0.5)^4 - 0.3,
      \end{aligned}$

\\ & $x_1$: \textit{age},  $x_2$: \textit{date of diagnosis}, $x_5$: \textit{if in age group within 71 to 95 years old} \\
      
\hline
 & 

$\begin{aligned}[t]
        f(\mathbf{x}; \hat{\bm{\Phi}}) & =  0.2\text{sin}(1.4x_1 + 6.8) - 0.2\text{sin}(1.6x_2 + 5.2) -0.6\text{tanh}(1.2x_3 + 1.3)- 0.8\text{sin}(1.2x_4 - 2.6)
      \end{aligned}$ 

 \\

FRTCS& $\begin{aligned}[t]
         \,\,- 0.2\text{sin}(1.8x_5 - 10) + 0.2(x_6+1)^2 -0.5\text{tanh}(2.3x_7 + 2) + 0.4\text{tanh}(1x_8 + 0.2) + 0.6(x_{12} + 0.2)^4 - 1.2,
      \end{aligned}$ 

\\ & $x_1$: \textit{age}, $x_2$: \textit{sbp0},   $x_3$: \textit{dbp0}  $x_4$: \textit{sbp1}, $x_5$: \textit{dbp1}, $x_6$: \textit{sbp2},   $x_7$: \textit{dbp2},  $x_8$: \textit{date0},   $x_{12}$: \textit{antihyp1}  \\
      
\hline
\end{tabular}
\label{tab:symbo_summary}
\end{table*}

\begin{figure}[!t]\centering
\subfigure[Impacts of order setting in spline function]{
\centering
\includegraphics[width=3.2in]{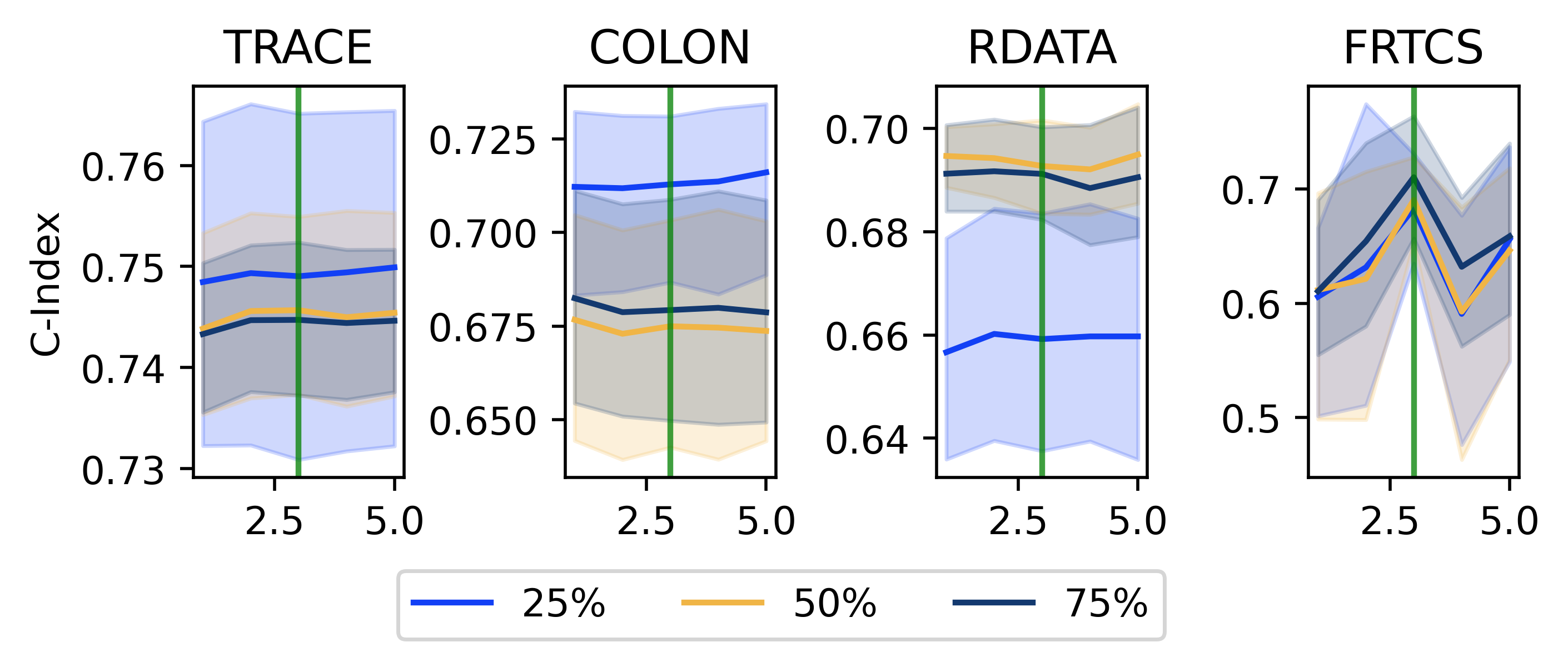}
\label{fig:ablation_order}
}%
\hfil
\subfigure[Impacts of $\gamma$ setting in loss function]{
\centering
\includegraphics[width=3.2in]{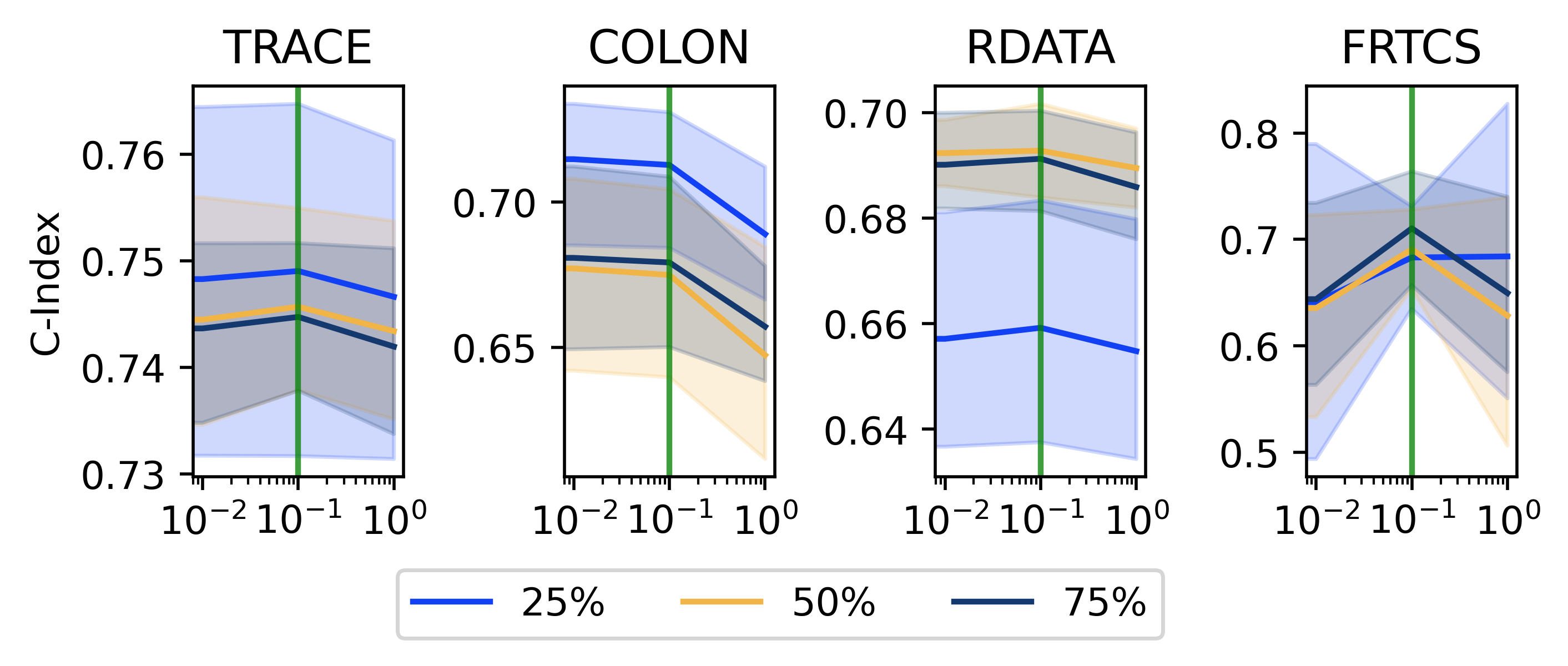}
\label{fig:ablation_lamb}
}%
\caption{\textbf{Ablation study on model configurations on real-world datasets}, including impacts of (a) order setting in spline function and (b) regularization term in loss function.}
\label{fig:ablation}
\end{figure}

\subsubsection{Experiments with Real-World Data}

The results from experiments on real-world datasets, namely TRACE, COLON, RDATA, and FRTCS, are also summarized in Table \ref{tab:summary} and Figure \ref{fig:box}. These results demonstrate the competitive performance of GCPH. While DCPH achieves the best performance on TRACE, with GCPH as the second-best, GCPH performs within half of the best results in the remaining datasets. Additionally, GCPH provides symbolic functions that illustrate the relationships between covariates and risk of patients.

Figure \ref{fig:trace_sym} presents the symbolic functions estimated for each variable in TRACE. The variables include the measure of heart pumping effect (\textit{wmi}, where 2 is normal and 0 is worst), age, sex (1 if female), clinical heart pump failure (\textit{chf}, 1 if present), diabetes (1 if present), and ventricular fibrillation (\textit{vf}, 1 if present). 
The estimated functions indicate that the presence of clinical heart pump failure, diabetes, or ventricular fibrillation increases patient risk. Conversely, a higher measure of heart pumping effect is associated with lower risk, and older patients face higher risk. Gender appears to have less impact on risk, as shown in Figure \ref{fig:trace_sym_3}.

We summarize the symbolic non-linear log-risk functions approximated by GCPH for various public benchmarks in Table \ref{tab:symbo_summary}. For simplicity, we present the functions with only two decimal places, excluding activation functions with minimal weights to highlight important features.

\subsection{Ablation Study}

We conducted an ablation study to evaluate the impact of different model configurations in our experiments.

\subsubsection{Impact of Order Setting} 
Figure \ref{fig:ablation_order} shows the model performance using varying spline function orders, ranging from 1 to 5, with $\gamma=0.1$. The results indicate that the model performance is generally less sensitive to changes in order. However, significant impacts were observed on the FRTCS dataset, with the setting of $K=3$ yielding optimal performance in our experiments.

\subsubsection{Impact of Regularization Loss} 
We also evaluated the impact of the regularization term in the loss function by varying $\gamma$. As shown in Figure \ref{fig:ablation_lamb}, the regularization term has a more pronounced effect across different datasets compared to the order setting. Overall, our choice of $\gamma=0.1$ provided the best performance among the configurations tested.

\section{Conclusion}

In this paper, we propose an extended CPH model that incorporates a symbolic non-linear log-risk function. This function is approximated using the KAN model, allowing for an effective symbolic representation of the relationships between covariates and survival outcomes. We integrate the log-partial likelihood function into the loss function to update the GCPH model, enabling it to perform survival analysis. Compared to extended CPH models using MLPs, which often involve many trainable parameters and lack interpretability, our model offers a transparent and streamlined formulation of the log-risk function. This approach provides valuable insights into how different variables influence survival outcomes in an interpretable manner.

\bibliographystyle{unsrt}
\bibliography{coxkan}

\end{document}